\documentclass{article}

\usepackage{amsmath,amsfonts,bm}

\def\eqref#1{equation~\ref{#1}}

\def\1{\bm{1}}

\DeclareMathAlphabet{\mathsfit}{\encodingdefault}{\sfdefault}{m}{sl}
\SetMathAlphabet{\mathsfit}{bold}{\encodingdefault}{\sfdefault}{bx}{n}

\def\gF{{\mathcal{F}}}

\def\gT{{\mathcal{T}}}

\def\gV{{\mathcal{V}}}

\def\gX{{\mathcal{X}}}
\def\gY{{\mathcal{Y}}}

\newcommand{\R}{\mathbb{R}}

\usepackage[pagenumbers]{cvpr}

\usepackage{url}

\usepackage{amssymb}
\usepackage{booktabs}
\usepackage{comment}
\usepackage{graphicx}
\usepackage{lipsum}
\usepackage{multirow}
\usepackage{subcaption}
\usepackage[dvipsnames]{xcolor}

\usepackage{colortbl} 

\newcommand\blfootnote[1]{%
  \begin{NoHyper}%
  \renewcommand\thefootnote{}\footnote{#1}%
  \addtocounter{footnote}{-1}%
  \end{NoHyper}%
}

\definecolor{cvprblue}{rgb}{0.21,0.49,0.74}
\usepackage[pagebackref,breaklinks,colorlinks,allcolors=cvprblue]{hyperref}

\title{Visual-Word Tokenizer: Beyond Fixed Sets of Tokens in Vision Transformers}
\author{
    \textbf{Leonidas Gee}\textsuperscript{\normalfont1*} \quad \textbf{Wing Yan Li}\textsuperscript{\normalfont2} \quad \textbf{Viktoriia Sharmanska}\textsuperscript{\normalfont1} \quad \textbf{Novi Quadrianto}\textsuperscript{\normalfont1,3,4}
    \\
    \textsuperscript{1}Predictive Analytics Lab, University of Sussex, UK \quad \textsuperscript{2}University of Surrey, UK \\ \textsuperscript{3}Basque Center for Applied Mathematics, Spain \quad \textsuperscript{4}Monash University, Indonesia
}
\date{}

\begin{document}
\maketitle
\blfootnote{* Corresponding author: jg717@sussex.ac.uk.}

\begin{center}
    \captionsetup{type=figure}
    \includegraphics[scale=0.8]{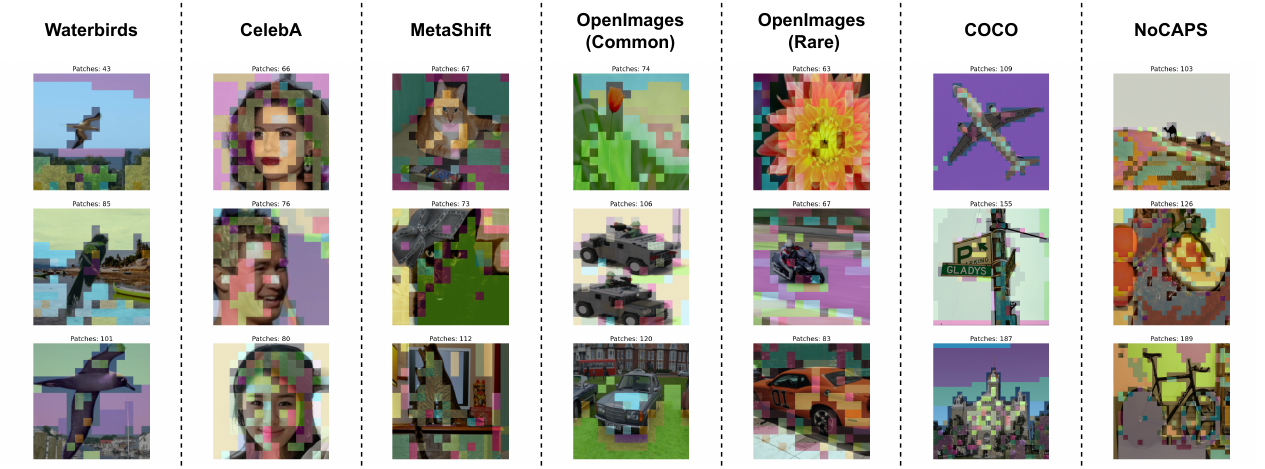}
    \captionof{figure}{
        Visualization of the \emph{inter}-image approach by the Visual-Word Tokenizer ($\gT_{inter}^{100}$ model). 
        Patches that are matched with one another are indicated by identical colors. 
        Higher patch matching is exhibited with the background rather than foreground object across the datasets. 
        Patch matching serves as a rudimentary form of image segmentation by grouping similar non-adjacent visual concepts during inference.
    }
    \label{fig:match_inter}
\end{center}

\begin{abstract}
    \small
    \emph{
        The cost of deploying vision transformers increasingly represents a barrier to wider industrial adoption. 
        Existing compression techniques require additional end-to-end fine-tuning or incur a significant drawback to energy efficiency, making them ill-suited for online (real-time) inference, where a prediction is made on any new input as it comes in. 
        We introduce the \textbf{Visual-Word Tokenizer} (VWT), a training-free method for reducing energy costs while retaining performance. 
        The VWT groups visual subwords (image patches) that are frequently used into visual words, while infrequent ones remain intact. 
        To do so, \emph{intra}-image or \emph{inter}-image statistics are leveraged to identify similar visual concepts for sequence compression. 
        Experimentally, we demonstrate a reduction in energy consumed of up to 47\%. 
        Comparative approaches of 8-bit quantization and token merging can lead to significantly increased energy costs (up to 500\% or more). 
        Our results indicate that VWTs are well-suited for efficient online inference with a marginal compromise on performance. 
        The experimental code for our paper is also made publicly available\footnote{\url{https://github.com/wearepal/visual-word-tokenizer}}.
    }
\end{abstract}

\section{Introduction}

In recent years, deep learning has seen continuous integration into a variety of systems worldwide. 
From coding to gaming, neural networks are increasingly deployed in online scenarios where asynchronous requests are processed in real-time. 
However, due to the size and complexity of modern architectures, such models are costly to run in practice. 
To address this, various methods have been proposed to improve model efficiency such as Knowledge Distillation \citep{hinton2015distilling}, Pruning \citep{han2015learning, michel2019sixteen}, and Quantization \citep{dettmers2022gpt3}. 
Many of these methods either require end-to-end fine-tuning to recover performance or significantly reduce energy efficiency. 
In the field of Natural Language Processing (NLP), there is a growing trend towards improving efficiency via tokenization \citep{gee2022fast, gee2023multi, dagan2024getting, yamaguchi2024empirical, minixhofer2024zero}. 
Newer large language models (LLMs) \citep{team2024gemma, dubey2024llama} exhibit a noticeably larger vocabulary than their earlier counterparts \citep{devlin2018bert, radford2019language}, thereby producing shorter sequences across various distributions. 
For computer vision, increasing interest is placed on reducing the cost of deploying the vision transformer (ViT) \citep{dosovitskiy2020image}. 
As image encoders in larger vision-language systems, ViTs are used to process images as fixed sets of tokens. 
Similar to downsampling in convolutional neural networks, most research \citep{kong2022spvit, liang2022not, rao2021dynamicvit, marin2021token, bolya2022token, bian2023multi, kim2024token} has focused on merging and/or pruning tokens in the intermediate layers to reduce computational overhead. 
Given the analogous architecture of the transformer across modalities, our work looks instead at the idea of tokenization for efficiency by splitting an image into variable sets of tokens.

To introduce variability, we draw inspiration from subword tokenization algorithms \citep{gage1994new, sennrich2016neural} used in NLP, which follow the principle that common words should remain intact while infrequent ones are broken down into meaningful subword units. 
Instead of a top-down approach -- splitting words into subwords, our work for image data takes a bottom-up approach by grouping visual subwords (image patches) into visual words. 
We also twist the underlying principle: frequently used patches should be grouped as they are more likely to describe common features, while infrequent ones remain intact as they might carry task-relevant information. 
We propose two procedures to capture this principle. 
The first is an \emph{intra}-image approach where patches with the lowest pixel variance within each image are grouped as they typically represent uniform areas (e.g., backgrounds). 
The second is an \emph{inter}-image approach where basic features across multiple images such as colors or edges are discovered as visual words. 
Image patches are then grouped based on the similarity of these basic characteristics. 
Crucially, patches that have distinct characteristics (i.e., high dissimilarity with any visual word) remain intact and form separate visual subwords.

\section{Related Work}

\paragraph{Efficient ViTs.}

Most works for improving the efficiency of ViTs have focused on reducing tokens in the intermediate layers by leveraging importance scores. 
In \citet{liang2022not, xu2022evo, kong2022spvit, bian2023multi}, redundancy is addressed by fusing tokens. 
Both \citet{rao2021dynamicvit} and \citet{tang2022patch} opt to prune such tokens instead. 
Recent efforts \citep{cao2023pumer, chen2023diffrate, bonnaerens2023learned, kim2024token} attempt to combine the benefits of merging and pruning. 
In \citet{tran2024accelerating}, an additional metric termed the energy score is used to better identify redundancy. 
Uniquely, \citet{fayyaz2022adaptive} use inverse transform sampling to select important tokens. 
Most relevant to our work are \citet{marin2021token} and \citet{bolya2022token}. 
The former assigns tokens to centroids via clustering, while the latter progressively merges tokens layer-by-layer in a training-free manner\footnote{Unlike \citet{bolya2022token}, we do not include \citet{marin2021token} as one of our baselines due to a lack of code release.}.

\begin{figure}[!t]
        \centering
        \resizebox{\textwidth}{!}{%
            \includegraphics{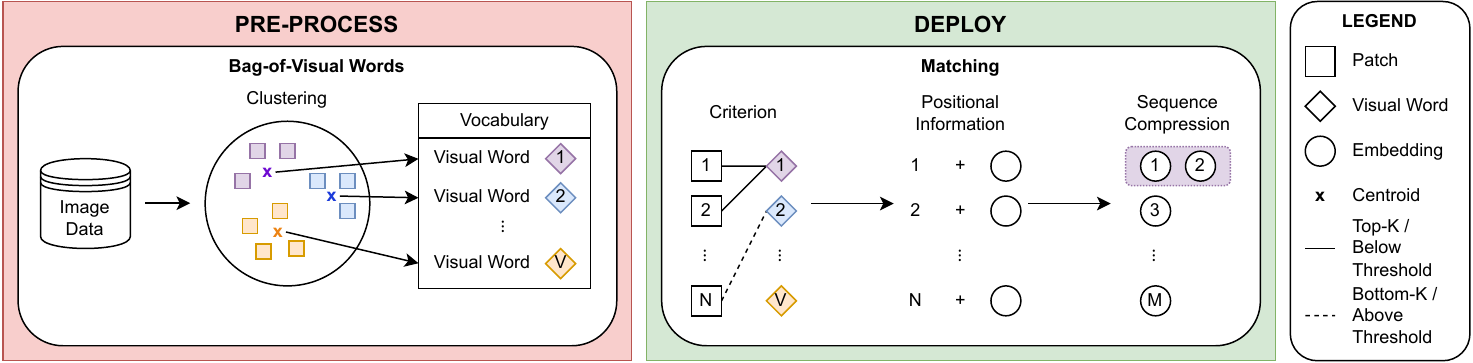}
        }
        \caption{
            Overview of the \textbf{Visual-Word Tokenizer} (VWT). 
            An \emph{intra}-image approach (\textsc{deploy} only). 
            During inference, the pixel variance of the patches is computed with the top-k lowest values being masked. 
            The masked tokens are dropped after positional information is added. 
            An \emph{inter}-image approach (\textsc{pre-process} \& \textsc{deploy}). 
            First, a Bag-of-Visual Words is formed by clustering patches in the pixel space (\textsc{pre-process}). 
            Then, during inference, the minimum pairwise cosine distance between the patches and visual words is computed with values above the threshold being masked (\textsc{deploy}). 
            The unmasked tokens are averaged based on their grouping to the same visual word after positional information is added.
        }
    \label{fig:schema}
\end{figure}

\paragraph{Specialized Tokenizers.}

Our method also takes inspiration from efficient inference in NLP. 
Increasingly, the tokenizer's vocabulary is specialized to reduce the input token length. 
In \citet{gee2022fast}, domain adaptation of the tokenizer ensures fewer subword or character tokens are produced. 
\citet{gee2023multi} followed up by introducing n-grams for tokenization beyond the word-level boundary. 
In \citet{dagan2024getting}, tokenizer specialization is also shown to accelerate the task of code generation with modern LLMs. 
Meanwhile, \citet{yamaguchi2024empirical} analyzed the effectiveness of various vocabulary adaptation techniques for efficient cross-lingual inference. 
Recently, \citet{minixhofer2024zero} leveraged hypernetworks for zero-shot tokenizer transfer of newly domain-adapted vocabularies.

\paragraph{Vector Quantization.}

The idea of discretizing continuous distributions has been explored in many works, most recently for image generation. 
\citet{yang2007evaluating} leveraged clustering for learning a codebook that maps keypoint descriptors to discrete visual words. 
In \citet{wu2020visual} and \citet{bao2021beit}, discretization is applied as part of the modelling for ViTs. 
\citet{van2017neural} learned discrete image representations by introducing the \emph{Vector Quantised-Variational Autoencoder} (VQ-VAE) approach. 
\citet{esser2021taming} and \citet{yu2021vector} further improved upon the VQ-VAE by combining the expressiveness of transformers with an adversarial framework. 
Increasingly, vision-language models paired with codebooks conduct image synthesis autoregressively \citep{ramesh2021zero, yu2022scaling, lu2022unified, lu2024unified, team2023gemini, team2024chameleon, sun2024autoregressive}. 
Lastly, \citet{yang2022visual} tackled disentangled representation learning and scene decomposition by tokenizing images into separate visual concepts.

\section{Visual-Word Tokenizer}

\begin{table}[!t]
    \centering
    \resizebox{\textwidth}{!}{%
        \begin{tabular}{r|c|c|c|cc|cc|cc}
            \toprule

            \multirow{2}{*}{\textbf{Dataset}} & \multirow{2}{*}{\textbf{Model}} & \multirow{2}{*}{\textbf{$Base$}} & \multirow{2}{*}{\textbf{$\gT_{intra}^{0.5}$}} & \multicolumn{2}{c|}{\textbf{$\gT_{inter}^{100}$}} & \multicolumn{2}{c|}{\textbf{$\gT_{inter}^{1000}$}} & \multicolumn{2}{c}{\textbf{$\gT_{inter}^{10000}$}}    \\

            & & & & \textbf{In-Domain} & \textbf{ImageNet}        & \textbf{In-Domain} & \textbf{ImageNet}        & \textbf{In-Domain} & \textbf{ImageNet}  \\

            \midrule

            Waterbirds & \multirow{5}{*}{CLIP}      & \multirow{5}{*}{197} & \multirow{5}{*}{99} & 125 & 124     & 144 & 144     & 165 & 169 \\
            CelebA & & &                                                                         & 89  & 88      & 130 & 119     & 163 & 155 \\
            MetaShift & & &                                                                      & 112 & 109     & 136 & 135     & 162 & 164 \\
            OpenImages (Com.) & & &                                                              & -   & 114     & -   & 136     & -   & 162 \\
            OpenImages (Rare) & & &                                                              & -   & 110     & -   & 133     & -   & 160 \\

            \midrule

            COCO & \multirow{2}{*}{BLIP}        & \multirow{2}{*}{577} & \multirow{2}{*}{289} & 267 & 264     & 317 & 312     & 408 & 405 \\
            NoCaps & & &                                                                      & -   & 257     & -   & 307     & -   & 403 \\

            \bottomrule
        \end{tabular}
    }
    \caption{
        Token length per sample (including [CLS]). 
        $\gT_{inter}^{\gV}$ of varying pre-processing data and vocabulary sizes are shown. 
        VWTs significantly reduce the token length of $Base$, particularly for larger image sizes (384 $\times$ 384 on COCO and NoCaps). 
        Unlike text tokenizers, domain specialization (In-Domain) does not result in greater compression for the $inter$-image approach. 
        Smaller vocabularies produce shorter sequences whereas larger vocabularies result in patches being increasingly matched to separate visual words.
    }
    \label{tab:length}
\end{table}

\begin{table}[!t]
    \centering
        \begin{tabular}{r|c|c|cc|>{\columncolor{yellow!20}}c|>{\columncolor{yellow!20}}c>{\columncolor{yellow!20}}c>{\columncolor{yellow!20}}c}
            \toprule

            \textbf{Dataset} & \textbf{Model} & \textbf{$Base$} & \textbf{$Q_{8}$} & \textbf{$ToMe$} & \textbf{$\gT_{intra}^{0.5}$} & \textbf{$\gT_{inter}^{100}$} & \textbf{$\gT_{inter}^{1000}$} & \textbf{$\gT_{inter}^{10000}$} \\

            \midrule

            Waterbirds & \multirow{5}{*}{CLIP}      & 1.05  &   6.54  &   1.56  &   0.65  &   1.01  &   1.05  &   1.16      \\
            CelebA &                                & 1.05  &   6.66  &   1.61  &   0.67  &   0.91  &   0.97  &   1.15      \\
            MetaShift &                             & 1.01  &   6.62  &   1.57  &   0.68  &   0.99  &   1.04  &   1.12      \\
            OpenImages (Com.) &                     & 1.19  &   6.90  &   1.65  &   0.68  &   1.03  &   1.09  &   1.27      \\
            OpenImages (Rare) &                     & 1.17  &   6.47  &   1.60  &   0.72  &   1.01  &   1.11  &   1.19      \\

            \midrule

            COCO & \multirow{2}{*}{BLIP}            & 2.39  &   4.66  &   2.22  &   1.38  &   1.26  &   1.42  &   1.77      \\
            NoCaps &                                & 2.30  &   4.72  &   2.25  &   1.41  &   1.34  &   1.44  &   1.79      \\

            \bottomrule
        \end{tabular}

    \caption{
        Energy (joule) consumed per sample. 
        The power and runtime values that were used to compute the energy consumed are provided in Table \ref{tab:power_runtime}.
        Compared to $Base$, \colorbox{yellow!20}{VWTs} reduce energy usage by up to 43\% with $\gT_{intra}^{0.5}$ and 47\% with $\gT_{inter}^{100}$. 
        Both $Q_{8}$ and $ToMe$ significantly increase energy costs, particularly on datasets with smaller image sizes (224 $\times$ 224 on Waterbirds, CelebA, MetaShift, OpenImages). 
        For the \emph{inter}-image approach, efficiency improvements are highest with larger image sizes (384 $\times$ 384 on COCO and NoCaps) and decreases naturally with larger vocabularies due to smaller compression.
    }
    \label{tab:efficiency}
\end{table}

In ViTs, tokenization is a process that splits an image into patches (tokens) which are then projected into a series of embeddings. 
The number of patches is typically fixed (e.g., $197$) based on the choice of architecture. 
We seek to split an image into variable length inputs instead (i.e., certain images will use $80$ tokens, while others a $100$). 
This variability is induced by conventional text tokenizers \citep{gage1994new, sennrich2016neural, wu2016google, kudo2018sentencepiece} for model efficiency. 
We propose to achieve this via visual words that group patches (visual subwords) based on some commonality in the \textbf{pixel space}. 
These visual words capture frequently used patches while retaining infrequent ones as is. 
A simple yet effective grouping can be done using either a criterion that looks at statistics of only one image (an \emph{intra}-image approach) or across many images (an \emph{inter}-image approach). 
Figure \ref{fig:schema} summarizes the \textbf{Visual-Word Tokenizer} (VWT).

\begin{table}[!t]
    \begin{subtable}{\textwidth}
    \centering
    \resizebox{\textwidth}{!}{%
        \begin{tabular}{c|cc|cc|cc|cc}
            \toprule

            \multirow{2}{*}{\textbf{Model}} & \multicolumn{2}{c|}{\textbf{Waterbirds}} & \multicolumn{2}{c|}{\textbf{CelebA}} & \multicolumn{2}{c}{\textbf{MetaShift}} & \multicolumn{2}{c}{\textbf{OpenImages}}  \\

            & \textbf{Average} $\uparrow$ & \textbf{Worst} $\uparrow$       & \textbf{Average} $\uparrow$ & \textbf{Worst} $\uparrow$       & \textbf{Average} $\uparrow$ & \textbf{Worst} $\uparrow$       & \textbf{Common} $\uparrow$ & \textbf{Rare} $\uparrow$     \\

            \midrule

            $Base$      & 79.06 & 21.86     & 89.61 & 47.21     & 95.31 & 87.69     & 70.48 & 63.36     \\

            \midrule

            $Q_{8}$     & 79.94 & 24.25     & 89.73 & 48.19     & 95.23 & 88.21     & 70.48 & 63.19     \\
            $ToMe$      & 79.80 & 27.05     & 89.25 & 28.52     & 94.24 & 87.37     & 65.69 & 60.05     \\

            \midrule

            \rowcolor{yellow!20}
            $\gT_{intra}^{0.5}$         & 75.56 & 31.00        & 89.27 & 50.93      & 92.94 & 86.15     & 65.52 & 59.73     \\[1mm]

            \midrule

            \rowcolor{yellow!20}
            $\gT_{inter}^{100}$         & 78.41 & 26.01        & 90.04 & 53.15      & 90.24 & 84.28     & 62.90 & 58.10     \\[1mm]

            \rowcolor{yellow!20}
            $\gT_{inter}^{1000}$        & 79.19 & 23.83        & 90.79 & 47.96      & 93.94 & 86.67     & 66.15 & 60.56     \\[1mm]

            \rowcolor{yellow!20}
            $\gT_{inter}^{10000}$       & 79.68 & 22.90        & 90.12 & 46.85      & 94.81 & 86.15     & 69.03 & 62.50     \\
    
            \bottomrule
        \end{tabular}
    }
    \caption{CLIP (image classification and subgroup robustness)}
    \end{subtable}

    \hspace{1em}

    \begin{subtable}{\textwidth}
    \centering
    \resizebox{\textwidth}{!}{%
        \begin{tabular}{c|cc|cccccccc}
            \toprule

            \multirow{3}{*}{\textbf{Model}}        & \multicolumn{2}{c|}{\textbf{COCO}}        & \multicolumn{8}{c}{\textbf{NoCaps}}    \\

            & \multicolumn{2}{c|}{\textbf{Karpathy} $\uparrow$} & \multicolumn{2}{c}{\textbf{In-Domain} $\uparrow$}       & \multicolumn{2}{c}{\textbf{Near-Domain} $\uparrow$}     & \multicolumn{2}{c}{\textbf{Out-of-Domain} $\uparrow$}      & \multicolumn{2}{c}{\textbf{Overall} $\uparrow$}    \\

            & BLEU@4 & CIDEr        & CIDEr & SPICE     & CIDEr & SPICE     & CIDEr & SPICE     & CIDEr & SPICE    \\

            \midrule

            $Base$      & 34.00 & 107.35        & 103.57 & 14.60        & 98.87  & 14.11        & 92.96 & 13.30     & 98.34 & 14.02     \\

            \midrule

            $Q_{8}$     & 33.85 & 106.87        & 103.86 & 14.53        & 100.07 & 14.18        & 92.72 & 13.49     & 99.12 & 14.10     \\
            $ToMe$      & 30.02 & 94.40         & 97.35  & 14.42        & 90.74  & 13.39        & 84.22 & 12.69     & 90.37 & 13.40     \\

            \midrule

            \rowcolor{yellow!20}
            $\gT_{intra}^{0.5}$         & 32.80 & 104.37        & 99.86  & 14.20        & 94.21 & 13.73     & 89.52 & 13.12     & 94.07 & 13.68     \\[1mm]

            \midrule

            \rowcolor{yellow!20}
            $\gT_{inter}^{100}$         & 31.10 & 97.70         & 95.58  & 13.78        & 89.62 & 13.11     & 82.42 & 12.44     & 89.02 & 13.08     \\[1mm]

            \rowcolor{yellow!20}
            $\gT_{inter}^{1000}$        & 32.12 & 101.79        & 98.19  & 13.95        & 93.85 & 13.49     & 86.03 & 12.67     & 92.89 & 13.39     \\[1mm]

            \rowcolor{yellow!20}
            $\gT_{inter}^{10000}$       & 33.18 & 105.08        & 102.84 & 14.59        & 98.03 & 13.95     & 91.52 & 13.14     & 97.40 & 13.88     \\
    
            \bottomrule
        \end{tabular}
    }
    \caption{BLIP (image captioning)}
    \end{subtable}

    \caption{
        Zero-shot (training-free) image classification (CLIP), subgroup robustness (CLIP) and captioning (BLIP). 
        $Q_{8}$ displays the lowest degradation in performance relative to $Base$, but suffers from a significant increase in energy used as shown in Table \ref{tab:efficiency}. 
        The VWTs are competitive with $ToMe$, particularly for image captioning on COCO and NoCaps. 
        Both $Q_{8}$ and the \colorbox{yellow!20}{VWTs} are shown to improve the worst-group accuracies on Waterbirds and CelebA. 
        For the latter, the $intra$-image approach works best on MetaShift and OpenImages while the $inter$-image approach is more suitable on the remaining datasets.
        The lower performance of the $inter$-image approach on MetaShift and OpenImages may stem from the lack of uniform backgrounds (e.g., skies in Waterbirds) that increasingly shifts compression to the foreground object.
    }
    \label{tab:performance}
\end{table}

\subsection{An \emph{intra}-image approach}

The pixel variance of the image patches is the simplest criterion that can be used for grouping. 
In Figure~\ref{fig:schema}, this approach only utilizes the \textsc{deploy} step by grouping the top-k patches with the lowest pixel variance while leaving the rest intact. 
To compress the grouped tokens, we opt to drop them as they tend to exhibit excessive homogeneity. 
We do not include the [CLS] token for dropping. 
This approach is inspired by \citet{minderer2024scaling} which aimed to reduce the training cost of OWLv2, an open-vocabulary object detector. 
Dropping patches with the lowest pixel variance removes padding areas and uniform backgrounds from the input mosaics of raw images, thereby increasing training efficiency.

\subsection{An \emph{inter}-image approach}

Inspired by text tokenizers and codebooks, we propose a variable-length approach that statistically discovers visual words across many images. 
The tokenizer consists of two steps: \textsc{Pre-Process} and \textsc{Deploy}.

\begin{figure}[!t]
    \centering
    \resizebox{\textwidth}{!}{%
        \includegraphics{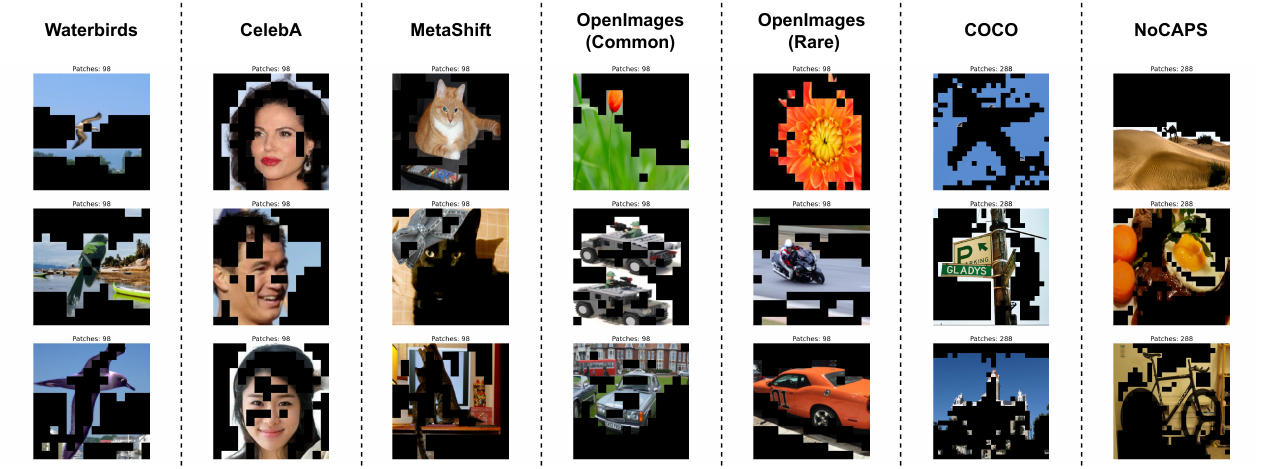}
    }
    \caption{
        Visualization of the \emph{intra}-image approach by the Visual-Word Tokenizer ($\gT_{intra}^{0.5}$ model). 
        Patches with the lowest pixel variance that are dropped are indicated in black. 
        In most cases, the dropped patches correspond to the uniform background which is uninformative. 
        However, in a few cases, the patches of the foreground object are dropped (e.g., airplane in first image for COCO) which is undesirable.
    }
    \label{fig:match_intra}
\end{figure}

\begin{figure}[!t]
    \centering
    \resizebox{\textwidth}{!}{%
        \includegraphics{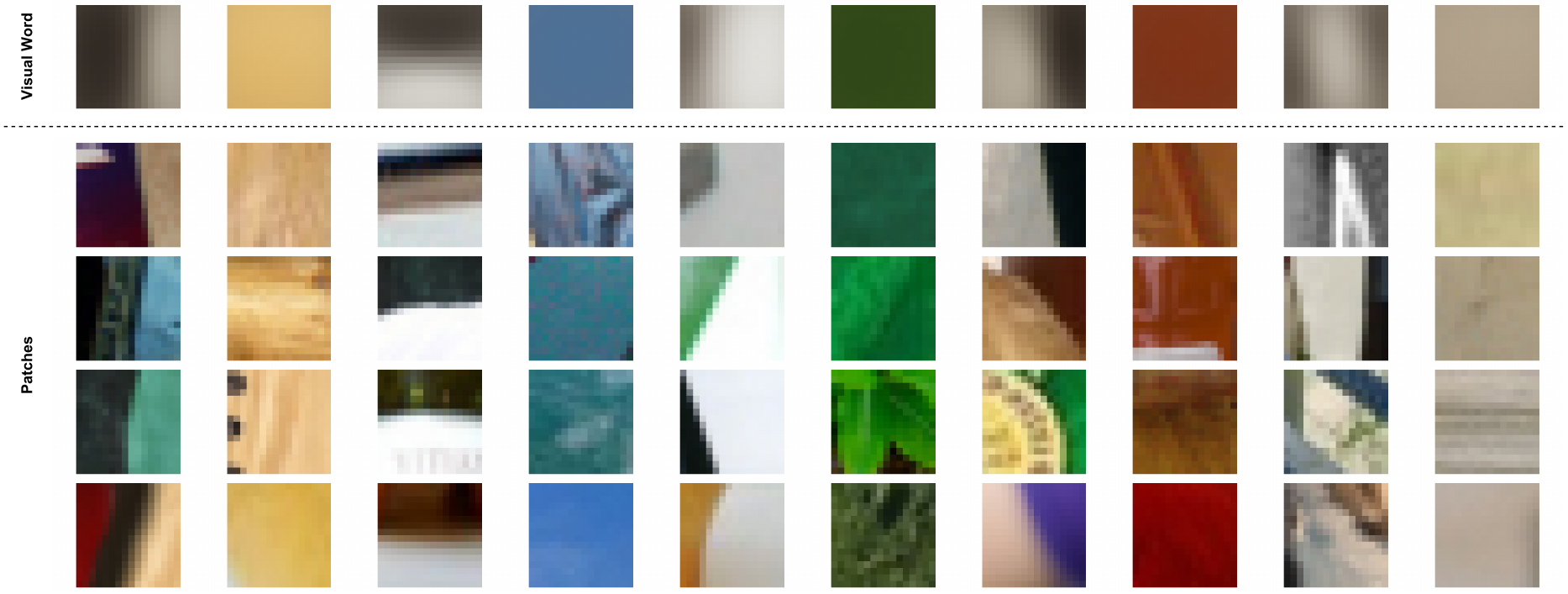}
    }
    \caption{
        Visualization of the vocabulary of $\gT_{inter}^{100}$. 
        Patches from ImageNet-1K are matched to the closest visual word using the Euclidean distance. 
        Visual words are shown to depict basic features such as colors or edges. 
        Each visual word is an average representation of patches that belong to the matched cluster.
    }
    \label{fig:words}
\end{figure}

\paragraph{\textsc{Pre-Process}.}

The Bag-of-Visual Words (BoVW) is a popular method for modeling images via discrete representations. 
In \citet{yang2007evaluating}, k-means clustering is applied to keypoint descriptors from SIFT \citep{lowe2004distinctive} to learn a fixed set of centroids. 
These centroids represent the vocabulary to which multiple descriptors are mapped in a process termed \emph{Vector Quantization} (VQ). 
In our method, we adopt a variation of this framework by building the BoVW using patches within the pixel space. 
Our design choice is motivated by two main factors. 
First, we find keypoint descriptors to be costly for inference. 
In each forward pass, computing keypoints for each image would significantly increase runtime. 
Second, in our early experimentation, we observed that patches in the embedding space have little similarity to one another. 
Such issues were also described by \citet{bolya2022token}, thus leading to their use of attention scores instead. 
Further justification for leveraging the pixel space is provided in Section \ref{sec:embedding}.

Given a dataset, we first patchify the images using the same patch size as the image encoder (e.g., 16 for ViT-B/16). 
We then cluster the patches via k-means to form the BoVW of a given vocabulary size, where each centroid represents a visual word. 
Patchification is done via basic tensor operations\footnote{We decompose the image into smaller tensors given the patch size while avoiding any linear transformations.} and not the pre-trained convolutional layer of the ViT to avoid projection into embeddings. 
We also execute this process in batches using the MiniBatchKMeans\footnote{from sklearn.cluster import MiniBatchKMeans} algorithm due to memory constraints. 
Note that MiniBatchKMeans uses the Euclidean distance by design. 
Since clustering is done in the pixel space, the BoVW may be reused by other ViTs with the same patch size regardless of model type.

\begin{figure}[!t]
    \centering
    \resizebox{\textwidth}{!}{%
        \includegraphics{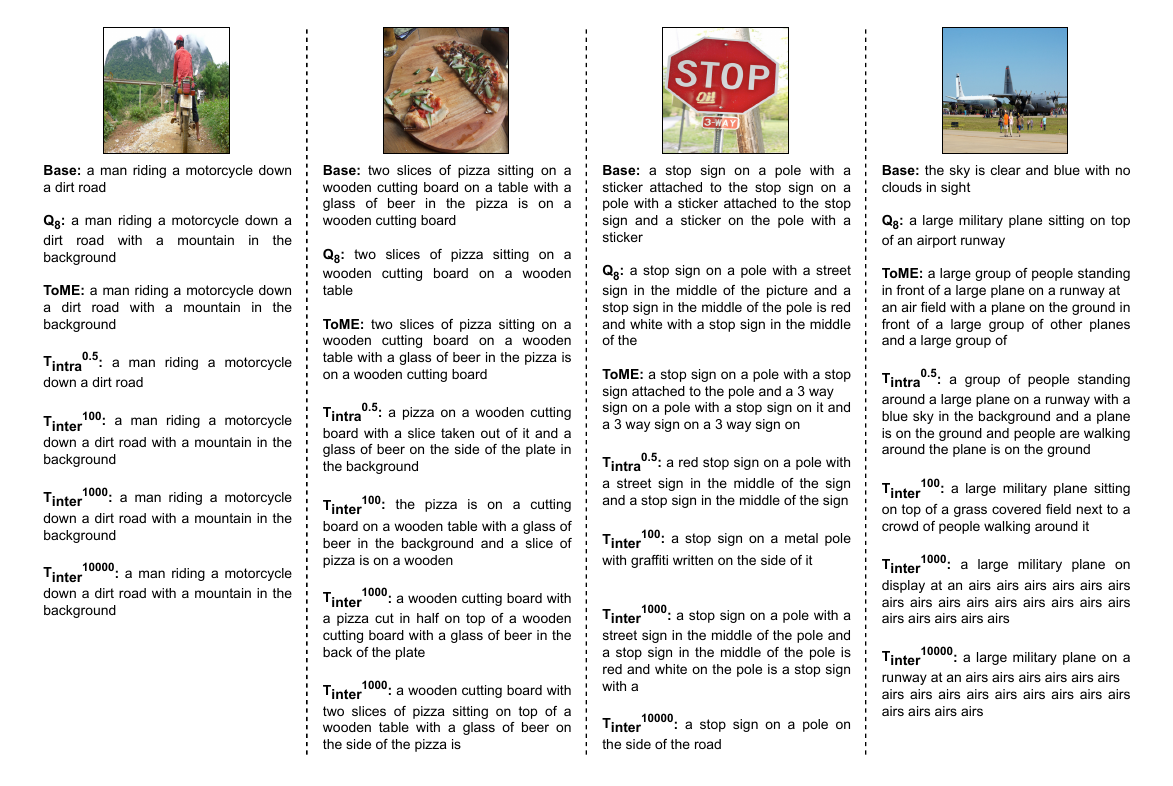}
    }
    \caption{
        Visualization of long-form captioning on COCO. 
        Longer captions are generated via a length penalty of 2.0 and a maximum length of 40. 
        Interestingly, the smaller vocabulary of $\gT_{inter}^{100}$ possesses higher descriptiveness and coherence than $\gT_{intra}^{0.5}$, $\gT_{inter}^{1000}$, or $\gT_{inter}^{10000}$ despite its higher compression.
    }
    \label{fig:long}
\end{figure}

\paragraph{\textsc{Deploy}.}

Once the BoVW is formed, we turn towards the process of sequence compression. 
One way of leveraging the BoVW would be to merge similar patches in the pixel space before projecting them into embeddings. 
However, such a naive approach will significantly degrade performance as the initialization of a new linear layer for projection is required. 
To avoid this, we begin by patchifying and computing the pairwise cosine distance between the patches and BoVW. 
For each patch, we retain only the minimum distance. 
Unlike text, we are unable to obtain exact matches with images. 
As such, distances higher than a given threshold are masked out to ensure dissimilar patches are not merged.
At this point, we have determined the groupings of similar patches via their connections to the same visual words. 
We then apply the pre-trained convolutional layer of the ViT on the original image to patchify and project it into a series of embeddings. 
Before merging, we ensure that positional information is added to the embeddings as we found it to work better than adding them later. 
Lastly, we average the embeddings element-wise based on the earlier defined groupings. 
We do not include the [CLS] token for merging.

For the \emph{inter}-image approach, if batching instead of online inference is desired, the uniform requirement of tensors becomes a challenge. 
To maintain parallelism, any reduction has to be equal across samples. 
Due to the non-uniformity of tokenization, sequences have to be sequentially compressed before padding to the same length. 
We opt to append additional [PAD] tokens until the longest length within the batch is achieved. 
Similar to text transformers \citep{vaswani2017attention}, the attention scores are set to negative infinity before the softmax to nullify the influence of padding. 
We do not add positional information to the [PAD] tokens as extrapolating such information to non-uniform sequences will significantly worsen model efficiency.

\section{Experiments}

Consider a pre-trained image encoder \( f \in \gF \) with parameters \( \theta \in \Theta \) that transforms inputs \( x \in \gX \subseteq \R^d \) to encodings \( \hat{x} \in \hat{\gX} \subseteq \R^{\hat{d}} \). 
The encodings can then be mapped to labels $y \in \gY$ for some given task, be it classification or generation. 
More specifically, the ViT first transforms inputs \( x \) to tokens \( t_{1}, \ldots, t_{N} \) before further processing by the attention layers. 
The number of tokens \( N \) is a constant defined by \( (\frac{I}{P})^{2} \), where $I$ and $P$ are the image and patch sizes, respectively. 
Let \( \gT \) be the VWT associated with a vocabulary \( v \in \gV \), where \( v \) is a visual word learned from some dataset \( D \). 
The tokenizer transforms the input \( x \) into tokens \( t_{1}, \ldots, t_{M} \), where \( M \ll N \). 
In our experiments, we seek to analyze the effect of \( \gT \) for online inference. 
We focus on the zero-shot setting by eschewing any form of end-to-end fine-tuning for $f$.

\begin{table}[!t]
    \begin{subtable}{1\textwidth}
    \centering
    \resizebox{1\textwidth}{!}{%
        \begin{tabular}{c|cc|cc|cc|cc}
            \toprule

            \multirow{2}{*}{\textbf{Model}} & \multicolumn{2}{c|}{\textbf{Waterbirds}} & \multicolumn{2}{c|}{\textbf{CelebA}} & \multicolumn{2}{c}{\textbf{MetaShift}}      & \multicolumn{2}{c}{\textbf{OpenImages}}   \\

            & \textbf{Average} $\uparrow$ & \textbf{Worst} $\uparrow$      & \textbf{Average} $\uparrow$ & \textbf{Worst} $\uparrow$        & \textbf{Average} $\uparrow$ & \textbf{Worst} $\uparrow$    & \textbf{Common} $\uparrow$ & \textbf{Rare} $\uparrow$   \\

            \midrule

            $\gT_{intra}^{0.5} + Q_{8}$         & 76.31 & 33.65     & 89.43 & 52.41     & 93.14 & 87.78     & 65.29 & 59.27     \\[1mm]
            $\gT_{inter}^{100} + Q_{8}$         & 78.88 & 27.57     & 90.17 & 54.44     & 90.58 & 83.72     & 62.75 & 57.99     \\[1mm]
            $\gT_{inter}^{1000} + Q_{8}$        & 80.01 & 25.60     & 91.05 & 50.93     & 93.67 & 86.67     & 66.15 & 60.48     \\[1mm]
            $\gT_{inter}^{10000} + Q_{8}$       & 80.28 & 25.29     & 90.26 & 47.96     & 94.58 & 86.67     & 69.07 & 62.37     \\
    
            \bottomrule
        \end{tabular}
    }
    \caption{CLIP (image classification and subgroup robustness)}
    \end{subtable}

    \hspace{1em}

    \begin{subtable}{1\textwidth}
    \centering
    \resizebox{1\textwidth}{!}{%
        \begin{tabular}{c|cc|cccccccc}
            \toprule

            \multirow{3}{*}{\textbf{Model}}     & \multicolumn{2}{c|}{\textbf{COCO}}     & \multicolumn{8}{c}{\textbf{NoCaps}}  \\

            & \multicolumn{2}{c|}{\textbf{Karpathy} $\uparrow$} & \multicolumn{2}{c}{\textbf{In-Domain} $\uparrow$}       & \multicolumn{2}{c}{\textbf{Near-Domain} $\uparrow$}     & \multicolumn{2}{c}{\textbf{Out-of-Domain} $\uparrow$}      & \multicolumn{2}{c}{\textbf{Overall} $\uparrow$}    \\

            & BLEU@4 & CIDEr        & CIDEr & SPICE     & CIDEr & SPICE     & CIDEr & SPICE     & CIDEr & SPICE    \\

            \midrule

            $\gT_{intra}^{0.5} + Q_{8}$         & 32.45 & 103.18        & 101.98 & 14.46        & 94.93 & 13.74        & 87.13 & 12.78     & 94.36 & 13.66     \\[1mm]
            $\gT_{inter}^{100} + Q_{8}$         & 30.88 & 97.48         & 95.97  & 13.94        & 90.27 & 13.26        & 80.02 & 12.39     & 89.01 & 13.19     \\[1mm]
            $\gT_{inter}^{1000} + Q_{8}$        & 31.72 & 100.23        & 99.69  & 14.02        & 94.35 & 13.49        & 85.69 & 12.57     & 93.36 & 13.39     \\[1mm]
            $\gT_{inter}^{10000} + Q_{8}$       & 32.98 & 104.15        & 101.77 & 14.34        & 98.75 & 14.02        & 91.55 & 13.21     & 97.72 & 13.91     \\
    
            \bottomrule
        \end{tabular}
    }
    \caption{BLIP (image captioning)}
    \end{subtable}

    \caption{
        Zero-shot (training-free) image classification (CLIP), subgroup robustness (CLIP) and captioning (BLIP). 
        The VWTs are applied jointly with 8-bit quantization. 
        Performance is shown to be similar to those with VWTs only, thereby displaying the mutual compatibility of VWTs with other compression techniques.
    }
    \label{tab:vwt_quant}
\end{table}

\subsection{Datasets and Settings}

We conduct our analysis through the lens of (i) classification performance of visual recognition, (ii) subgroup robustness, and (iii) generative performance of visual captioning. 
For (i) and (ii), we utilize three publicly available datasets (Waterbirds \citep{wah2011caltech}, CelebA \citep{liu2015deep}, MetaShift \citep{liang2022metashift}) that are typical benchmarks in robustness and fairness research \citep{sagawa2019distributionally, liu2021just, yang2023change}. 
To perform (zero-shot) classification, we compute the cosine similarity between an image embedding and the following encoded text labels\textsuperscript{\ref{note:prefix}} for Waterbirds, CelebA, and MetaShift, respectively: \{'landbird', 'waterbird'\}, \{'non-blond', 'blond'\}, \{'dog', 'cat'\}. 
Further details on the defined subgroups are provided in Appendix \ref{sec:datasets}. 
For (i), we also conduct a large-scale evaluation on the OpenImages v6 dataset \citep{kuznetsova2020open}. 
Following \citet{huang2023open}, the test split is divided into \textit{common} and \textit{rare} subsets that consist of 57 224 and 21 991 images, respectively. 
The former has 214 classes, while the latter has 200. 
To perform (zero-shot) classification, we compute the cosine similarity between an image embedding and the encoded text label\footnote{\label{note:prefix}The prefix "a photo of a " is also added to encode each text label.}. 
For (iii), we utilize the Karpathy test split of COCO dataset \citep{lin2014microsoft} and a validation set of NoCaps dataset \citep{agrawal2019nocaps} following the settings in previous work \citep{li2022blip}. 
Finally, to study inference efficiency, we utilize all the datasets for the visual tasks (i)-(iii) described above in computing the energy (joule) consumed per sample. 
For calculating the power variable of energy, we call "pynvml.nvmlDeviceGetPowerUsage(handle)/1000" as done by Weights and Biases, which leverages the NVIDIA Management Library for monitoring and managing the NVIDIA GPUs. 
The experiments are also conducted in an isolated environment, thus ensuring the efficiency measurements are as accurate as possible. 
Note that we do not include the \textsc{Pre-Process} step of the \emph{inter}-image approach into our efficiency calculations, as this process is only done once and may be reused multiple times.

\subsection{Implementation Details} 

For image classification, we load the pre-trained CLIP \citep{radford2021learning} model from HuggingFace\footnote{https://huggingface.co/openai/clip-vit-base-patch16}. 
An image size of 224 $\times$ 224 is used with bilinear interpolation for CLIP. 
For image captioning, we load the pre-trained BLIP \citep{li2022blip} model from HuggingFace\footnote{https://huggingface.co/Salesforce/blip-image-captioning-base}. 
To perform zero-shot captioning, we use a beam size of 3 along with maximum and minimum lengths of 20 and 5, respectively. 
An image size of 384 $\times$ 384 is used with bicubic interpolation. 
Both CLIP and BLIP utilize the ViT-B/16 image encoder unless stated otherwise. 
Aside from the pre-trained model which we denote as $Base$, we also consider 8-bit quantization \citep{dettmers2022gpt3} and token merging \citep{bolya2022token} as additional baselines. 
We denote the former as $Q_{8}$ and the latter as $ToMe$. 
Following \citet{bolya2022token}, we utilize a reduction per layer for $ToMe$ of 13 with CLIP and 23 with BLIP due to their respective input image sizes. 
For the VWTs, we set the top-k of the \emph{intra}-image approach to 50\% of the total number of patches which we denote as $\gT_{intra}^{0.5}$ as we found it to work best. 
Additional dropping ratios are explored in Table \ref{tab:drop} of Appendix \ref{sec:ratios}. 
For the \emph{inter}-image approach, we set the threshold to 0.1 unless stated otherwise and denote it as $\gT_{inter}^{\gV}$, where $\gV$ is the size of the vocabulary. 
Lastly, our experiments are conducted using a single NVIDIA A100 GPU. 
Since our focus is on the \textbf{online setting} (real-time), we set the batch size to 1 unless stated otherwise.

\begin{table}[!t]
    \centering
    \resizebox{\textwidth}{!}{%
        \begin{tabular}{r|c|cc|cc|cc}
            \toprule

            \multirow{2}{*}{\textbf{Dataset}} & \multirow{2}{*}{\textbf{Model}} & \multicolumn{2}{c|}{\textbf{$\gT_{inter}^{100}$}} & \multicolumn{2}{c|}{\textbf{$\gT_{inter}^{1000}$}} & \multicolumn{2}{c}{\textbf{$\gT_{inter}^{10000}$}}    \\

            & & \textbf{In-Domain} & \textbf{ImageNet}        & \textbf{In-Domain} & \textbf{ImageNet}        & \textbf{In-Domain} & \textbf{ImageNet}  \\

            \midrule

            Waterbirds & \multirow{5}{*}{CLIP}      & 180 & 179     & 179 & 179     & 182 & 182     \\
            CelebA &                                & 153 & 154     & 157 & 154     & 170 & 165     \\
            MetaShift &                             & 176 & 173     & 175 & 174     & 180 & 180     \\
            OpenImages (Com.) &                     & -   & 171     & -   & 171     & -   & 176     \\
            OpenImages (Rare) &                     & -   & 169     & -   & 169     & -   & 174     \\

            \midrule

            COCO & \multirow{2}{*}{BLIP}            & 405 & 406     & 409 & 408     & 442 & 440     \\
            NoCaps &                                & -   & 394     & -   & 396     & -   & 431     \\

            \bottomrule
        \end{tabular}
    }
    \caption{
        Ablation of token length per sample (including [CLS]). 
        $\gT_{inter}^{\gV}$ of varying pre-processing data and vocabulary sizes are shown. 
        Visual words for the \emph{inter}-image approach are formed in the embedding space from patches after the pre-trained convolution layer of CLIP or BLIP. 
        Unlike Table \ref{tab:length}, poor compression is seen due to dissimilarity between the patches and new visual words during inference.
    }
    \label{tab:embed}
\end{table}

\subsection{Experimental Results}

\paragraph{VWTs and Inference Efficiency.}

We begin by analyzing the effects of VWTs on token length to understand how the choice of pre-processing data and vocabulary size affects the degree of compression. 
Table \ref{tab:length} shows the token length per sample (including [CLS]) on different datasets. 
First, we observe a significant reduction in token length by the VWTs relative to $Base$. 
This reduction is most apparent for datasets with larger image sizes such as COCO and NoCaps. 
On the former, the token length is reduced by up to 54\% with $\gT_{inter}^{100}$. 
Second, the token lengths induced by $\gT_{inter}^{\gV}$ are not equal unlike $\gT_{intra}^{0.5}$. 
We compare $\gT_{inter}^{\gV}$ pre-processed on the in-domain dataset and ImageNet-1K \citep{deng2009imagenet}. 
The in-domain dataset is represented by the training split if available. 
On text data, in-domain tokenizers \citep{gee2022fast, gee2023multi, dagan2024getting, yamaguchi2024empirical, minixhofer2024zero} have been shown to produce shorter sequences by specializing the vocabulary on the given distribution. 
Interestingly, we observe no such effect with image data as seen by the similar lengths between In-Domain and ImageNet-1K. 
Only on CelebA, do we see a slightly greater reduction with $\gT_{inter}^{1000}$ and $\gT_{inter}^{10000}$ pre-processed on ImageNet-1K. 
Third, unlike text tokenizers, decreasing compression is seen as vocabulary size increases. 
With text, larger vocabularies ensure that more tokens are kept as words rather than subwords. 
We posit that an increasing number of patches are matched to separate visual words, thus lowering the overall compression. 
Note that henceforth, we use ImageNet-1K as our pre-processing data for the \emph{inter}-image approach.

\begin{table}[!t]
    \begin{subtable}{\textwidth}
    \centering

        \begin{tabular}{r|c|c|c|cc|cc|cc}
            \toprule

            \textbf{Dataset} & \textbf{Model} & \textbf{$\gT_{inter}^{100}$} & \textbf{$\gT_{inter}^{1000}$} & \textbf{$\gT_{inter}^{10000}$}   \\

            \midrule

            Waterbirds & \multirow{5}{*}{CLIP}      & \multirow{5}{*}{88} & \multirow{5}{*}{179} & \multirow{5}{*}{195}   \\
            CelebA & & & &  \\
            MetaShift & & & &  \\
            OpenImages (Com.) & & & &  \\
            OpenImages (Rare) & & & &  \\

            \midrule

            COCO & \multirow{2}{*}{BLIP}            & \multirow{2}{*}{104} & \multirow{2}{*}{439} & \multirow{2}{*}{561}   \\
            NoCaps & & & &  \\

            \bottomrule
        \end{tabular}

    \caption{Token Length}
    \end{subtable}

    \hspace{1em}

    \begin{subtable}{\textwidth}
    \centering
    \resizebox{\textwidth}{!}{%
        \begin{tabular}{c|cc|cc|cc|cc}
            \toprule

            \multirow{2}{*}{\textbf{Model}} & \multicolumn{2}{c|}{\textbf{Waterbirds}} & \multicolumn{2}{c|}{\textbf{CelebA}} & \multicolumn{2}{c}{\textbf{MetaShift}} & \multicolumn{2}{c}{\textbf{Overall} $\uparrow$} \\

            & \textbf{Average} $\uparrow$ & \textbf{Worst} $\uparrow$      & \textbf{Average} $\uparrow$ & \textbf{Worst} $\uparrow$        & \textbf{Average} $\uparrow$ & \textbf{Worst} $\uparrow$   & \textbf{Common} $\uparrow$ & \textbf{Rare} $\uparrow$   \\

            \midrule

            $\gT_{inter}^{100}$         & 66.60 & 9.61      & 90.24 & 56.48     & 76.16 & 61.43     & 41.06 & 39.84     \\[1mm]
            $\gT_{inter}^{1000}$        & 78.10 & 22.74     & 90.25 & 47.96     & 94.20 & 87.18     & 70.29 & 63.48     \\[1mm]
            $\gT_{inter}^{10000}$       & 78.98 & 21.96     & 89.61 & 47.46     & 95.27 & 87.18     & 70.52 & 63.40     \\
    
            \bottomrule
        \end{tabular}
    }
    \caption{CLIP (image classification and subgroup robustness)}
    \end{subtable}

    \hspace{1em}

    \begin{subtable}{\textwidth}
    \centering
    \resizebox{\textwidth}{!}{%
        \begin{tabular}{c|cc|cccccccc}
            \toprule

            \multirow{3}{*}{\textbf{Model}}        & \multicolumn{2}{c|}{\textbf{COCO}}        & \multicolumn{8}{c}{\textbf{NoCaps}}    \\

            & \multicolumn{2}{c|}{\textbf{Karpathy} $\uparrow$} & \multicolumn{2}{c}{\textbf{In-Domain} $\uparrow$}       & \multicolumn{2}{c}{\textbf{Near-Domain} $\uparrow$}     & \multicolumn{2}{c}{\textbf{Out-of-Domain} $\uparrow$}      & \multicolumn{2}{c}{\textbf{Overall} $\uparrow$}    \\

            & BLEU@4 & CIDEr        & CIDEr & SPICE     & CIDEr & SPICE     & CIDEr & SPICE     & CIDEr & SPICE    \\

            \midrule

            $\gT_{inter}^{100}$       & 9.92  & 19.18         & 15.81  & 7.20         & 10.95 & 6.25      & 10.55 & 6.63      & 11.57 & 6.47      \\[1mm]
            $\gT_{inter}^{1000}$      & 32.06 & 100.52        & 96.87  & 14.04        & 91.83 & 13.47     & 89.52 & 12.96     & 92.09 & 13.45     \\[1mm]
            $\gT_{inter}^{10000}$     & 34.04 & 107.35        & 103.55 & 14.49        & 98.22 & 14.00     & 94.08 & 13.48     & 98.15 & 13.97     \\
    
            \bottomrule
        \end{tabular}
    }
    \caption{BLIP (image captioning)}
    \end{subtable}

    \caption{
        Ablation of token length (including [CLS]) and performance of the \emph{inter}-image approach with random merging. 
        The pairwise cosine distance is initialized by sampling from a uniform distribution of $[0, 2]$. 
        Average and worst-group accuracies degrade noticeably with $\gT_{inter}^{100}$ except for CelebA. 
        Performance does not change significantly for $\gT_{inter}^{1000}$ and $\gT_{inter}^{10000}$ from $Base$ as negligible compression occurs.
    }
    \label{tab:random_inter}
\end{table}

Having analyzed the effects on token length, we turn to the practical metrics of energy. 
In Table \ref{tab:efficiency}, we compare the efficiency of VWTs to $Base$, $Q_{8}$, and $ToMe$. 
Note that we compute both metrics using the image encoder of CLIP or BLIP only. 
First, we find the energy consumed by VWTs to be generally lower than $Base$ across the datasets. 
On OpenImages (Common) and COCO, this reduction is up to 43\% with $\gT_{intra}^{0.5}$ and 47\% with $\gT_{inter}^{100}$, respectively. 
Interestingly, the \emph{inter}-image approach appears to be most effective for datasets with larger image sizes (384 $\times$ 384) as shown with COCO and NoCaps rather than the remaining datasets with smaller image sizes (224 $\times$ 224). 
This efficiency also decreases as the vocabulary size increases due to smaller compression. 
For $Q_{8}$ and $ToMe$, we observe significant drawbacks in energy efficiency relative to $Base$. 
In the case of the former, the increase in energy consumed can reach as high as 500\% or more on Waterbirds, CelebA, and MetaShift. 
This can be explained by the significantly longer runtime of $Q_{8}$ as tensors need to be repeatedly quantized and dequantized. 
Meanwhile, $ToMe$ can increase energy consumption by up to 50\% or more. 
We have shown how VWTs are more suitable for online inference by improving energy efficiency, particularly for datasets with larger image sizes. 
As mentioned previously, we do not consider the \textsc{Pre-Process} step of the \emph{inter}-image approach in our efficiency calculations as it is only executed once and not repeated during inference. 
This is similar to how model efficiency for large language models is calculated by excluding the construction of the tokenizer's vocabulary via Byte-Pair Encoding. 
For reference, this step on the CPU takes approximately 30 minutes to an hour, depending on the vocabulary size\footnote{Note that this pre-processing is done on ImageNet-1K. Naturally, dataset size also affects the total duration.}.

\paragraph{VWTs and Visual Performance.}

Another important factor in compression is the effect on model performance. 
In Table \ref{tab:performance}, we tabulate the performance of image classification and captioning using CLIP and BLIP, respectively. 
For visual recognition and subgroup robustness, we analyze performance from an average and worst-group perspective as done by \citet{sagawa2019distributionally} and \citet{romiti2022realpatch}. 
To further validate performance, we report the mean Average Precision (mAP) on the large-scale OpenImages v6 dataset \citep{huang2023open}. 
For image captioning with BLIP, we evaluate our models following the setting in \citet{li2022blip} by using the BLEU, CIDEr, and SPICE metrics w.r.t. the ground truth captions.

First, we find the degradation in performance relative to $Base$ to be smallest for $Q_{8}$ across the datasets. 
However, this comes at the cost of a significant rise in energy consumed as shown in Table \ref{tab:efficiency}. 
We also observe that VWTs are competitive with $ToMe$, particularly on the image captioning tasks of COCO and NoCaps. 
Second, both $Q_{8}$ and the VWTs are shown to improve the subgroup robustness on Waterbirds and CelebA. 
The worst-group accuracy (WGA) with $\gT_{intra}^{0.5}$ increases by up to 29\% and 8\%, respectively. 
Only on MetaShift do we observe a lower WGA than $Q_{8}$ or $ToMe$. 
Like \citet{gee2023compressed}, we find compression to not always be harmful to subgroup robustness by improving the WGA at a small or negligible cost to overall performance. 
Third, the $intra$-image approach is seen to work best on MetaShift and OpenImages while the $inter$-image approach is more suitable on the remaining datasets. 
The average accuracy with $\gT_{inter}^{100}$ drops by up to 5\% and 7\% on MetaShift and OpenImages, respectively. 
We posit that the lower performance may stem from the lack of uniform backgrounds (e.g., skies in Waterbirds) that causes compression to increasingly shift to the foreground object instead as shown in Figure \ref{fig:match_inter}.

\begin{table}[!t]
    \centering
    \resizebox{\textwidth}{!}{%
        \begin{tabular}{c|ccc|ccc|ccc}
            \toprule

            \multirow{2}{*}{\textbf{Model}} & \multicolumn{3}{c|}{\textbf{Waterbirds}} & \multicolumn{3}{c|}{\textbf{CelebA}} & \multicolumn{3}{c}{\textbf{MetaShift}}  \\

            & \textbf{Length} & \textbf{Average} $\uparrow$ & \textbf{Worst} $\uparrow$       & \textbf{Length} & \textbf{Average} $\uparrow$ & \textbf{Worst} $\uparrow$       & \textbf{Length} & \textbf{Average} $\uparrow$ & \textbf{Worst} $\uparrow$     \\

            \midrule

            0.2     & 97 & 76.71 & 24.09     & 66 & 89.44 & 56.48     & 84 & 86.16 & 77.46     \\
            0.3     & 79 & 74.85 & 21.13     & 55 & 88.85 & 58.70     & 69 & 81.16 & 67.77     \\
            0.4     & 66 & 73.35 & 16.30     & 50 & 88.31 & 57.04     & 60 & 77.73 & 60.65     \\
            0.5     & 58 & 72.81 & 13.14     & 48 & 88.21 & 55.74     & 55 & 74.75 & 56.08     \\
    
            \bottomrule
        \end{tabular}
    }
    \caption{
        Ablation of token length (including [CLS]) and performance using $\gT_{inter}^{100}$ with varying similarity thresholds. 
        Higher thresholds for the \emph{inter}-image approach can reduce sequences to 48 tokens on CelebA. 
        Greater compression results in higher degradation in performance, particularly in subgroup robustness.
    }
    \label{tab:threshold}
\end{table}

\paragraph{Visualizing the VWTs.}

To better understand the VWTs, we visualize the patch dropping of $\gT_{intra}^{0.5}$ and patch matching of $\gT_{inter}^{100}$ in Figures \ref{fig:match_intra} and \ref{fig:match_inter}, respectively. 
For the latter, we highlight in identical colors the patches that are matched with one another. 
With $\gT_{intra}^{0.5}$, patches with the lowest pixel variance typically correspond to uninformative backgrounds. 
In most cases, dropping such patches continues to preserve the foreground object. 
With $\gT_{inter}^{100}$, we find that patches representing the background are more frequently grouped than those of the foreground object. 
On Waterbirds, the merging of background patches may explain the improved robustness in Table \ref{tab:performance} by mitigating spurious correlations with the background. 
On CelebA, $\gT_{inter}^{100}$ tends to avoid matching the eyes or mouths. 
We also observe that $\gT_{inter}^{100}$ may group similar but non-adjacent visual concepts. 
In certain examples of MetaShift and NoCaps, multiple cats and foods are matched together, respectively. 
Our analysis shows that patch matching serves as a rudimentary form of image segmentation by identifying similar visual concepts for compression.

In Figure \ref{fig:words}, we visualize the vocabulary of $\gT_{inter}^{100}$ to analyze the formation of the visual words. 
We show patches from ImageNet-1K (pre-processing data) that are matched to each visual word using the Euclidean distance. 
Since visual words are centroids, patches that are matched to the same visual word belong to the same cluster. 
We find that visual words depict basic features such as colors or edges. 
These features are also formed as an average representation of the patches that belong to each cluster. 
By representing basic features, visual words serve as effective anchor points to which patches can be matched to.

The generated captions by BLIP \citep{li2022blip} can be used as priors for further training of other models. 
As such, longer descriptive captions may be more desirable than the typical short captions associated with Internet data. 
In Figure \ref{fig:long}, we visualize the long-form captions on COCO. 
To enable longer generations, we set the length penalty to 2.0 and double the maximum length to 40. 
All other generation settings are kept the same. 
With longer captions, the generation may degenerate into unnecessary repetitions on certain samples. 
Interestingly, descriptiveness and coherence improves more with $\gT_{inter}^{100}$ than $\gT_{intra}^{0.5}$, $\gT_{inter}^{1000}$, or $\gT_{inter}^{10000}$ inspite of its higher sequence compression as seen on COCO in Table \ref{tab:length}.

\subsection{Ablation Studies}

\paragraph{VWTs with Quantization.}

We designed the VWT to be a training-free approach to efficient online inference. 
As such, we exclude any form of fine-tuning \citep{lin2024not,hao2024meft,simoulin2024memory} (e.g., knowledge distillation \citep{tian2019contrastive,han2024amd,son2021densely}) for performance recovery. 
However, should it be desired, our method may be used alongside network compression \citep{tian2019contrastive,han2024amd,son2021densely} (i.e., minimizing the Kullback–Leibler divergence between the softened outputs of the larger teacher and smaller student), memory-efficient fine-tuning \citep{hao2024meft,simoulin2024memory}, and other compression techniques (e.g., quantization) as the VWT only influences the initial input sequence to the vision transformer. 
In Table \ref{tab:vwt_quant}, we provide additional scores when VWTs are combined with 8-bit quantization. 
We find the performance to be similar to those in Table \ref{tab:performance}, thereby showing that VWTs are highly compatible with other compression approaches as mentioned above.

\begin{figure}[!t]
    \centering
    \resizebox{\textwidth}{!}{%
        \includegraphics{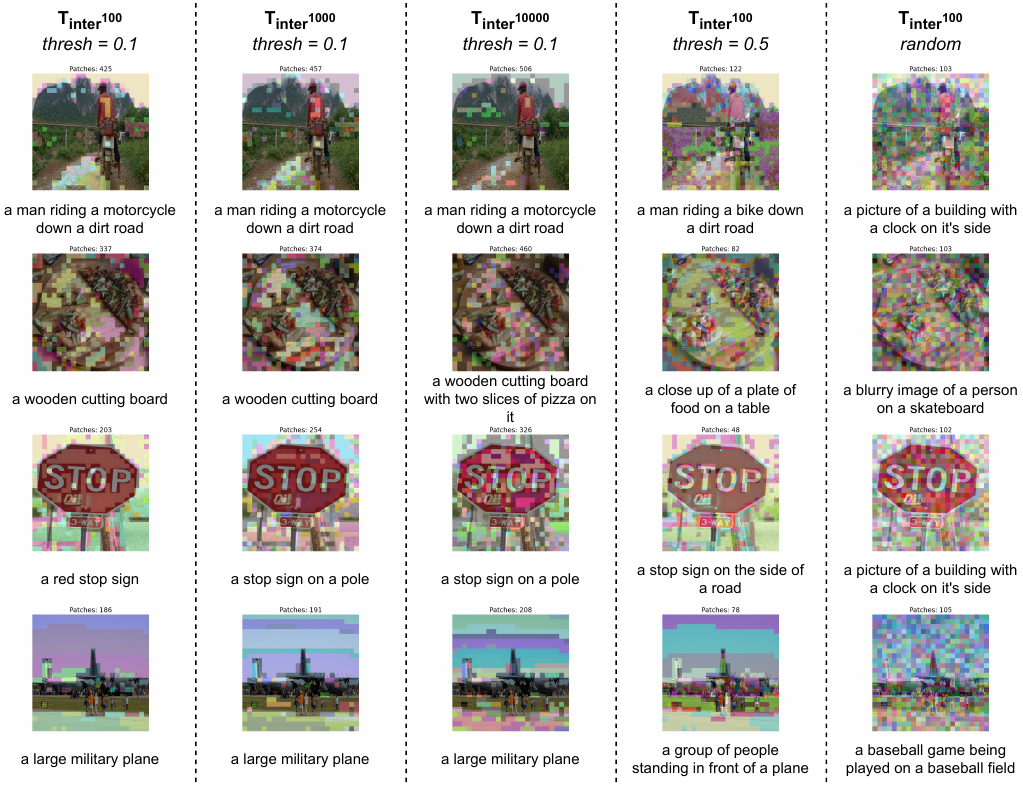}
    }
    \caption{
        Visualization of image captions on COCO by the \emph{inter}-image approach. 
        The generated captions are shown to deviate more when increasing the similarity threshold than when reducing the vocabulary size. 
        With random matching, the model begins to completely misunderstand the image.
    }
    \label{fig:cap_inter}
\end{figure}

\paragraph{Embeddings as Visual Words.}\label{sec:embedding}

In Table \ref{tab:embed}, we analyze compression by the \emph{inter}-image approach using visual words formed in the \textbf{embedding space}. 
Instead of initializing the vocabulary by clustering patches in the pixel space, we do so with patches after the pre-trained convolution layer of CLIP or BLIP. 
During inference, we match the patches after the pre-trained convolution layer with these new visual words. 
Compared to Table \ref{tab:length}, we observe a notable reduction in the compression irrespective of pre-processing data and vocabulary size. 
In \citet{tang2022patch}, embeddings are shown to become progressively more similar to one another due to self-attention. 
As such, it is unsurprising that matching patches and visual words in the initial embedding space (i.e., before any self-attention is applied) is found to be ineffective.

\paragraph{Raising the Similarity Threshold.}\label{sec:threshold}

We have shown how the \emph{inter}-image approach can improve the efficiency of online inference. 
To better understand its limitations, we ablate the similarity threshold of $\gT_{inter}^{100}$ by setting it to values of $\{0.2, 0.3, 0.4, 0.5\}$ in Table \ref{tab:threshold}. 
We seek to determine if exploiting higher thresholds for increased compression is a viable strategy. 
Naturally, we observe a reduction in performance as increasingly dissimilar patches are merged. 
For Waterbirds and MetaShift, the WGA degrades more significantly than the average, especially on the former. 
Interestingly, average accuracy remains relatively unchanged while WGA improves significantly on CelebA irrespective of similarity threshold. 
We posit that at higher thresholds, the \emph{inter}-image approach begins to merge core features that represent the foreground object, resulting in the reduced performance of Waterbirds and MetaShift during inference.

\paragraph{Random Merging of Tokens.}\label{sec:random}

In Table \ref{tab:random_inter}, we study the effects of randomly merging tokens to contrast the Bag-of-Visual-Words criterion for the \emph{inter}-image approach. 
We initialize the pairwise cosine distance by sampling from a uniform distribution of $[0, 2]$. 
First, we find the token length (including [CLS]) to differ noticeably from Table \ref{tab:length}. 
For $\gT_{inter}^{100}$, sequences are further reduced across the datasets. 
Conversely, $\gT_{inter}^{1000}$ and $\gT_{inter}^{10000}$ display little compression. 
Second, performance is shown to change from Table \ref{tab:performance}. 
We observe a significant degradation in average and worst-group accuracies with $\gT_{inter}^{100}$ on Waterbirds and MetaShift. 
For $\gT_{inter}^{1000}$ and $\gT_{inter}^{10000}$, performance does not shift much from $Base$ as barely any compression occurs. 
With random matching, the captions are shown to deviate completely from those of $Base$ in Figure \ref{fig:cap_inter}, thus further demonstrating that visual words are an effective criterion for grouping patches as shown in Figure \ref{fig:match_inter}.

\section{Conclusion}

In this work, we set out to define a training-free tokenization for ViTs that lowers the energy consumed while balancing costs to performance. 
In online scenarios, we have shown empirically that our \emph{intra}-image and \emph{inter}-image approaches are more suitable than 8-bit quantization and token merging for image classification and captioning. 
Visually, the criterion of the \emph{intra}-image approach typically corresponds to the background, while that of the \emph{inter}-image approach groups analogous visual concepts based on visual words that represent basic features. 
Concerning the choice between the \emph{intra}-image or \emph{inter}-image approach, when global information (e.g., image classification) is required, dropping tokens may be more advantageous by removing unnecessary noise from the visual input. 
On the other hand, when the task necessitates local information (e.g., long-form captioning), merging tokens may better preserve the visual concepts. 
This is also a limitation of our training-free approach that could restrict its viability in sensitive domains such as medical imaging or autonomous driving. 
For example, the image on the first row of Figure \ref{fig:cap_intra} of Appendix \ref{sec:ratios} shows that $\gT_{intra}^{0.5}$ removes the mountainous background, thus leading to the absence of "mountain" in the long-form caption of Figure \ref{fig:long}. 
As a future work, combining both approaches may maximize compression while safeguarding against the loss of critical information\footnote{For the current approach, a possible safeguard is to carefully tune the hyperparameters of the VWT (i.e., top-k or similarity threshold) to one's requirements such that foreground objects are not unintentionally affected.} such as by merging the mountains and dropping the bushes.


\section*{Acknowledgements}

This research was supported by a European Research Council (ERC) Starting Grant for the project ``Bayesian Models and Algorithms for Fairness and Transparency'', funded under the European Union's Horizon 2020 Framework Programme (grant agreement no. 851538). 
Novi Quadrianto is also supported by the Basque Government through the BERC 2022-2025 program and by the Ministry of Science and Innovation: BCAM Severo Ochoa accreditation CEX2021-001142-S / MICIN/ AEI/ 10.13039/501100011033. 
Viktoriia Sharmanska is currently at Epic Games.


{
    \small
    \bibliographystyle{ieeenat_fullname}
    \bibliography{main}
}

\section{Further Details}

\subsection{Datasets}\label{sec:datasets}

Here, we detail the task of each dataset for subgroup robustness. 
In Table \ref{tab:datasets}, we also tabulate the labels and attributes that define each subgroup along with their sample sizes.

\paragraph{Waterbirds.}
Given an image of a bird, the task is to predict whether it is a waterbird or landbird \citep{wah2011caltech}. 
Following \citet{sagawa2019distributionally}, the attribute is the background that the bird is on. 
We use the same dataset splits as \citet{liu2021just}.

\paragraph{CelebA.}
Given an image of a person, the task is to predict whether the hair color is blond or not \citep{liu2015deep}. 
Following \citet{sagawa2019distributionally}, the attribute is the binary gender of the person. 
We use the same dataset splits as \citet{liu2021just}.

\paragraph{MetaShift.}
Given an image of an animal, the task is to predict whether it is a dog or cat \citep{liang2022metashift}. 
Following \citet{liang2022metashift}, the attribute is the environment that the dog or cat is in. 
We use the same dataset splits as \citet{yang2023change}.



\section{Supplementary Experiments}


\subsection{Experimental Results}

\paragraph{Additional Dropping Ratios.}\label{sec:ratios}

\begin{figure}[!ht]
    \centering
    \resizebox{1\textwidth}{!}{%
        \includegraphics{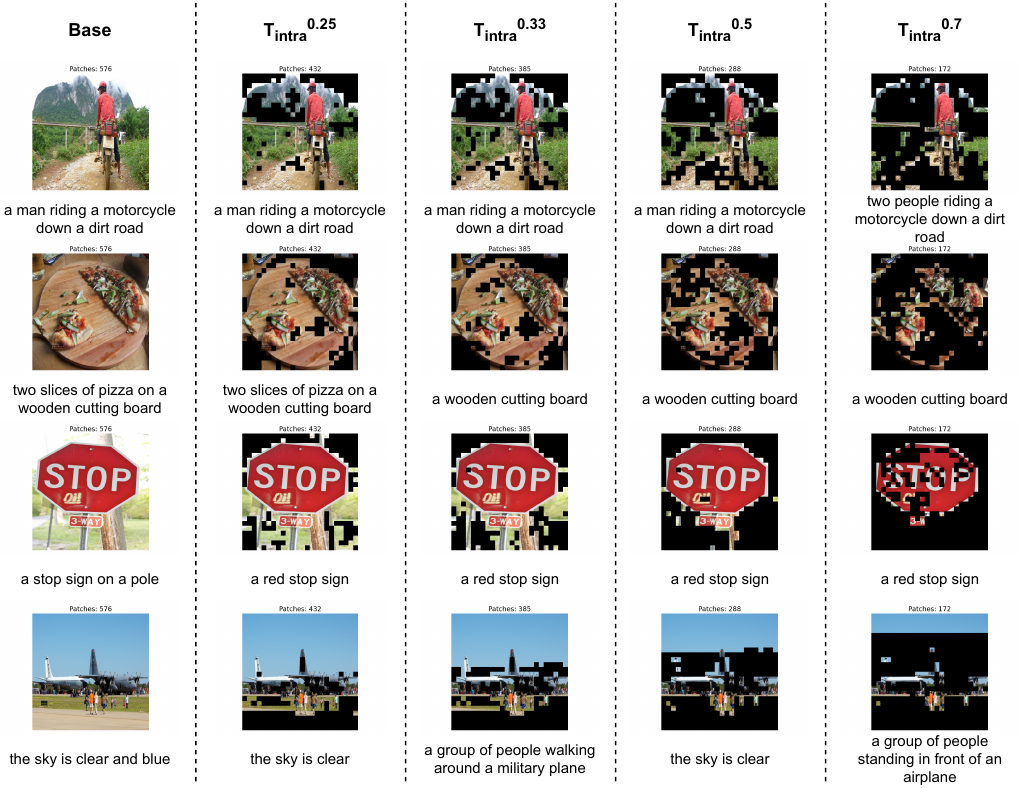}
    }
    \caption{
        Visualization of image captions on COCO by the \emph{intra}-image approach. 
        The generated captions to not deviate significantly from those of $Base$ with dropping ratios of up to 0.5.
    }
    \label{fig:cap_intra}
\end{figure}

We tabulate in Table \ref{tab:drop} the results of additional dropping ratios $\{0.25, 0.33, 0.7\}$ as used by \cite{minderer2024scaling}. 
We observe a natural degradation in performance as the ratio increases with a sharp drop at 0.7. 
Likewise, in Figure \ref{fig:cap_intra}, we find the image captions on COCO only begin to deviate significantly from $Base$ when dropping ratios are above 0.5.

\paragraph{Power and Runtime.}

In Table \ref{tab:power_runtime}, we provide the power (watt) and runtime (millisecond) that were used to calculate the energy (joule) consumed per sample in Table \ref{tab:efficiency}. 
Note that the runtime is first converted to seconds before taking its product with the power to produce the energy values.

\subsection{Ablation Studies}

\paragraph{Random Dropping of Tokens.}

In Table \ref{tab:random_intra}, we provide further analysis on the \emph{intra}-image approach by randomly dropping tokens. 
This is similar to \citet{simoulin2024memory} with the exclusion of any fine-tuning. 
First, unlike CelebA, we observe a noticeably degradation in worst-group accuracies on Waterbirds and MetaShift. 
We posit that this stems from the target label being spuriously correlated with the background (Waterbirds, MetaShift) and not gender (CelebA), which is distinctly separable from the foreground object as shown in Figure \ref{fig:match_intra}. 
Hence, random dropping is beneficial to the subgroup robustness of CelebA by reducing the gender features of the individuals. 
Second, we find performance (except $\gT_{intra}^{0.7}$) to be slightly lower or equivalent on OpenImages and the remaining datasets, respectively. 
We attribute this to cases where the \emph{intra}-image approach misidentifies the foreground object as irrelevant information in Figure \ref{fig:match_intra} with the removal of the “airplane” (top image) and “building” (bottom image). 
However, in general, the variance of the individual patches remains effective for robustly compressing redundant information within the image.

\paragraph{Fairness of Tokenization.}\label{sec:fairness}

Text tokenizers are known to induce unfairness between languages that raises compute costs, particularly for minority languages \citep{petrov2024language, ali2023tokenizer}. 
We seek to analyze if similar effects exist with VWTs. 
In Table \ref{tab:fairness}, we show the breakdown in token length (including [CLS]) and accuracy (w.r.t $Base$) by subgroup. 
First, we observe a notable difference in compression between the subgroups of Waterbirds. 
With $\gT_{inter}^{100}$, sequences might differ by up to 39 tokens as seen with subgroups 0 and 3. 
Smaller discrepancies are displayed on CelebA and MetaShift except for $\gT_{inter}^{100}$ on the former. 
Second, we find compression to not affect all subgroups equally. 
Accuracy improves on certain subgroups and degrades on others. 
Stronger compression does not necessarily correlate with a larger change in performance.

\paragraph{Sparsity of the Vocabulary.}\label{sec:sparsity}

\begin{figure}[!ht]
    \begin{subfigure}{\textwidth}
    \centering
        \includegraphics[width=0.3\textwidth]{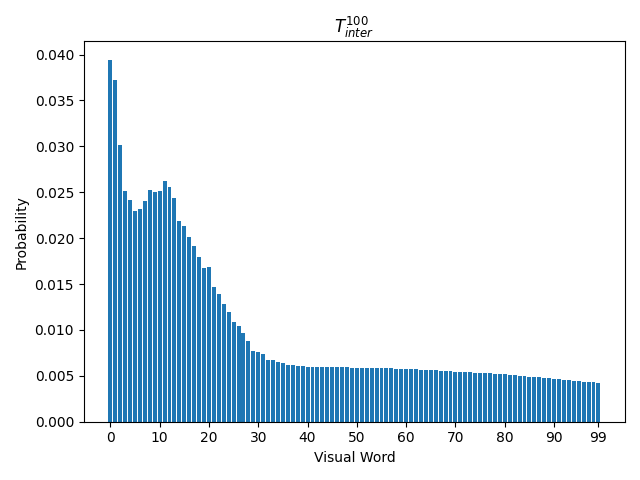}
        \includegraphics[width=0.3\textwidth]{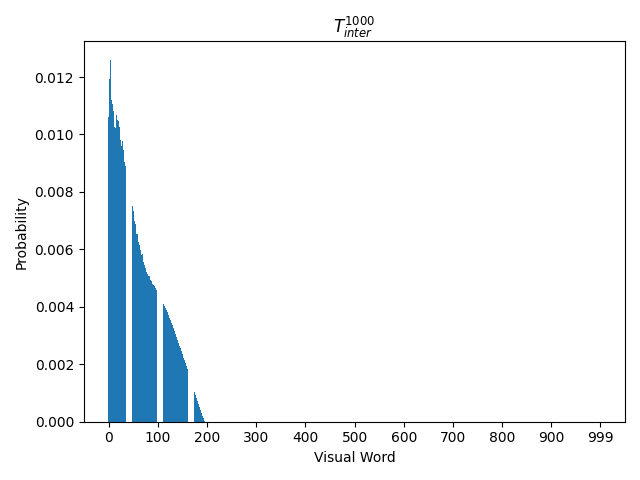}
        \includegraphics[width=0.3\textwidth]{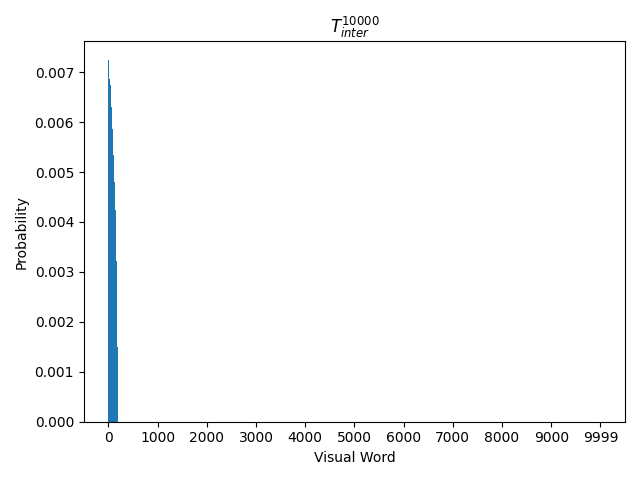}
        \caption{Waterbirds}
    \end{subfigure}

    \hspace{1em}

    \begin{subfigure}{\textwidth}
    \centering
        \includegraphics[width=0.3\textwidth]{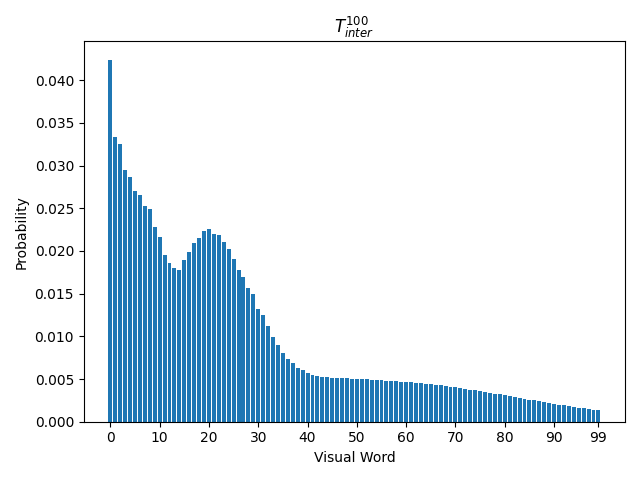}
        \includegraphics[width=0.3\textwidth]{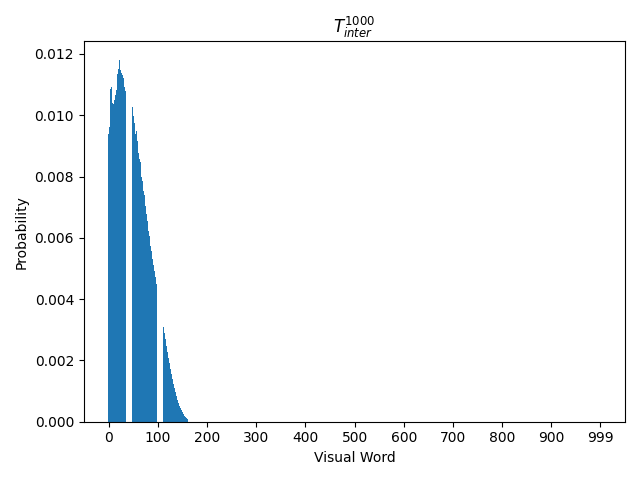}
        \includegraphics[width=0.3\textwidth]{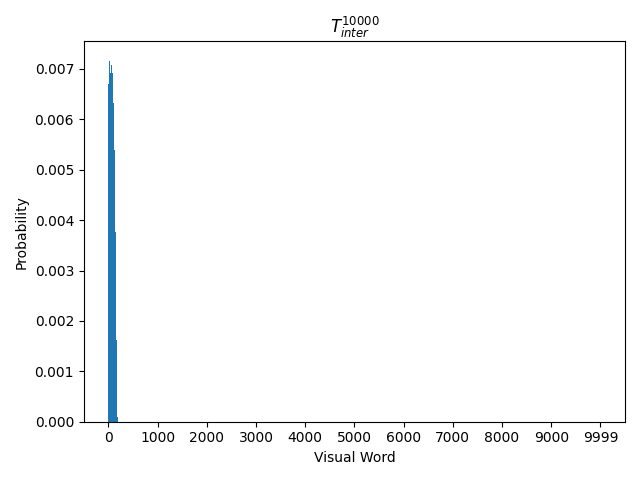}
        \caption{CelebA}
    \end{subfigure}

    \hspace{1em}

    \begin{subfigure}{\textwidth}
    \centering
        \includegraphics[width=0.3\textwidth]{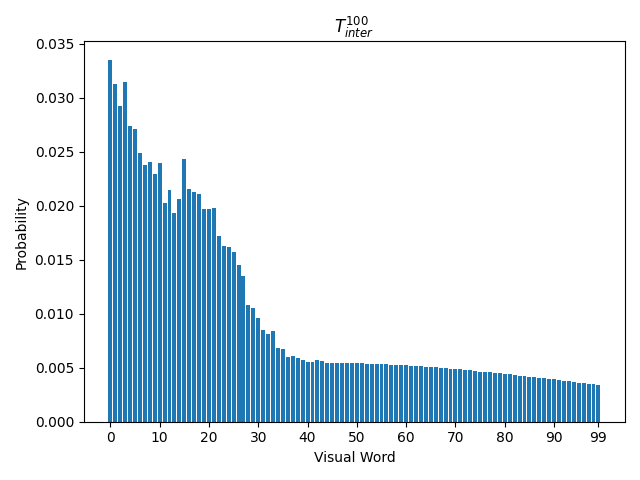}
        \includegraphics[width=0.3\textwidth]{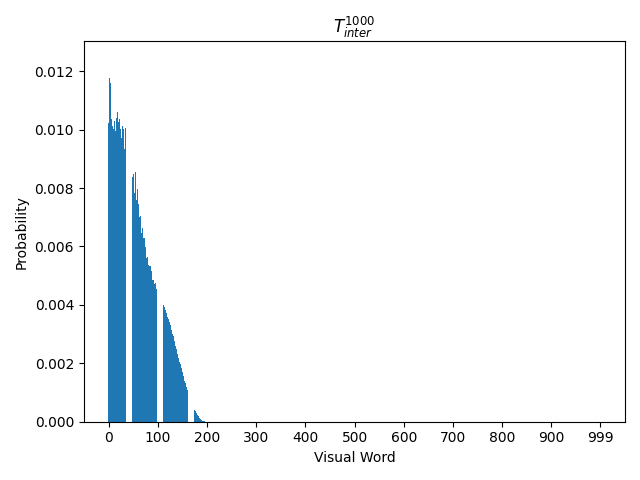}
        \includegraphics[width=0.3\textwidth]{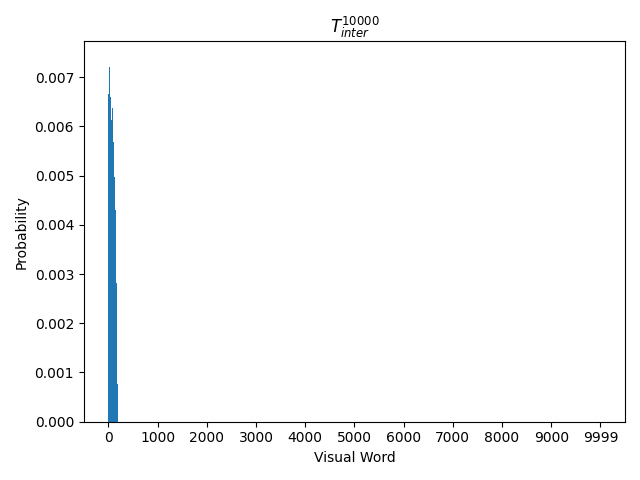}
        \caption{MetaShift}
    \end{subfigure}

    \caption{
        Probability distribution of the matched visual words. 
        $\gT_{inter}^{\gV}$ of varying vocabulary sizes are shown. 
        The probability distribution exhibits a large skew irrespective of the dataset as certain visual words are matched more frequently than others. 
        Larger vocabularies display greater sparsity as the many visual words that remain unmatched may be pruned for a more efficient vocabulary size. 
    }
    \label{fig:freq}
\end{figure}

To better understand the utilization of the visual words, we plot the probability distribution of the matches in Figure \ref{fig:freq}. 
Regardless of the dataset, we find that certain visual words are matched more frequently than others, thus leading to a large skew in the distributions. 
Greater sparsity is also displayed by larger vocabularies as many visual words remain unused across the datasets. 
As such, the pruning of unmatched visual words may be applied to achieve a more efficient vocabulary size after the \textsc{Pre-Process} step of the \emph{inter}-image approach.

\paragraph{VWTs with Batching.}\label{sec:batch}

\begin{figure}[!ht]
    \centering
    \resizebox{\textwidth}{!}{%
        \includegraphics{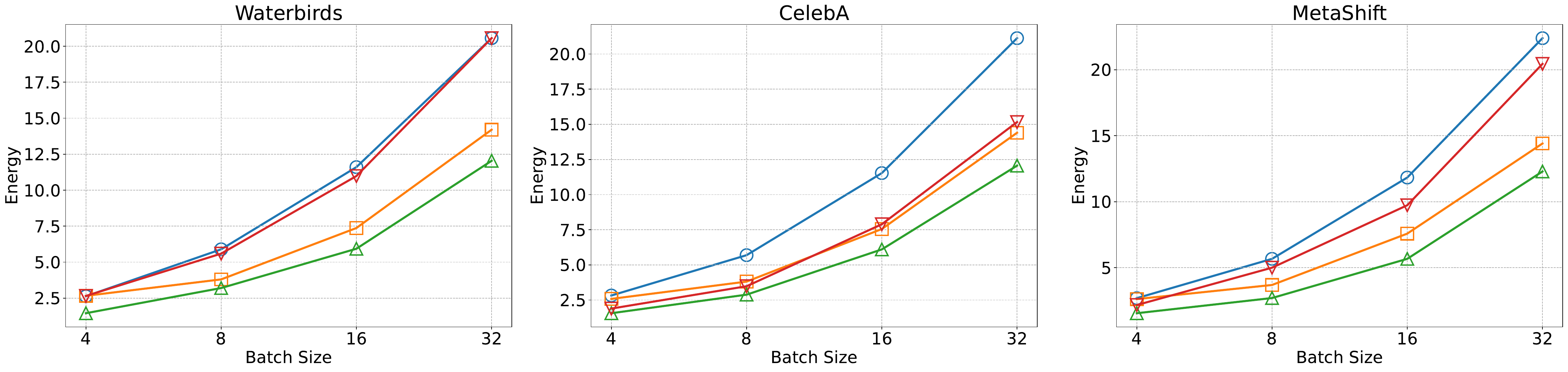}
    }

    \caption{
        Energy (joule) consumed with varying batch sizes. 
        The VWTs of $\gT_{intra}^{0.5}$ (\textcolor[HTML]{2ca02c}{$\bigtriangleup$}) and $\gT_{inter}^{100}$ (\textcolor[HTML]{d62728}{$\bigtriangledown$}) are compared to $Base$ (\textcolor[HTML]{1f77b4}{$\bigcirc$}) and $ToMe$ (\textcolor[HTML]{ff7f0e}{$\square$}). 
        The \emph{intra}-image approach displays the highest efficiency improvements followed by $ToMe$ and the \emph{inter}-image approach. 
        On Waterbirds, $\gT_{inter}^{100}$ does not display either a noticeable improvement or degradation in energy efficiency relative to $Base$.
    }
    \label{fig:batch}
\end{figure}

In Figure \ref{fig:batch}, we seek to better understand the effectiveness of VWTs for \textbf{offline inference} where batch sizes are greater than 1. 
Using batches of $\{4, 8, 16, 32\}$, we compare the energy (joule) consumed by the VWTs of $\gT_{intra}^{0.5}$ and $\gT_{inter}^{100}$ to $Base$ and $ToMe$. 
We find that as the batch size increases, both VWTs and $ToMe$ continue to generally improve energy efficiency across the datasets. 
This reduction in energy consumed is highest by $\gT_{intra}^{0.5}$ followed by $ToMe$ and $\gT_{inter}^{100}$. 
On Waterbirds, the \emph{inter}-image approach does not result in a noticeable improvement or degradation in energy consumed. 

\medskip

\begin{table}[!ht]
    \centering

        \begin{tabular}{ccccccc}
            \hline
            
            \textbf{Dataset}    & \textbf{Subgroup} & \textbf{Label}    & \textbf{Attribute}    & \textbf{Training} & \textbf{Validation}   & \textbf{Test} \\
    
            \hline
    
            \multirow{4}{*}{Waterbirds}     & 0     & 0 (landbird)              & 0 (on land)       & 3498  & 467   & 2255      \\
                                            & 1     & 0 (landbird)              & 1 (on water)      & 184   & 466   & 2255      \\
                                            & 2     & 1 (waterbird)             & 0 (on land)       & 56    & 133   & 642       \\
                                            & 3     & 1 (waterbird)             & 1 (on water)      & 1057  & 133   & 642       \\
    
            \hline
            
            \multirow{4}{*}{CelebA}         & 0     & 0 (non-blond)             & 0 (woman)         & 71629 & 8535  & 9767      \\
                                            & 1     & 0 (non-blond)             & 1 (man)           & 66874 & 8276  & 7535      \\
                                            & 2     & 1 (blond)                 & 0 (woman)         & 22880 & 2874  & 2480      \\
                                            & 3     & 1 (blond)                 & 1 (man)           & 1387  & 182   & 180       \\
    
            \hline
    
            \multirow{4}{*}{MetaShift}      & 0     & 0 (dog)                   & 0 (outdoor)       & 784   & 127   & 273       \\
                                            & 1     & 0 (dog)                   & 1 (indoor)        & 507   & 75    & 191       \\
                                            & 2     & 1 (cat)                   & 0 (outdoor)       & 196   & 33    & 65        \\
                                            & 3     & 1 (cat)                   & 1 (indoor)        & 789   & 114   & 345       \\
    
            \hline
        \end{tabular}

    \caption{
        Defined subgroups in Waterbirds, CelebA, and MetaShift.
    }
    \label{tab:datasets}
\end{table}

\begin{table}[!t]
    \begin{subtable}{1\textwidth}
    \centering

        \begin{tabular}{r|c|c|cc|c|ccc}
            \toprule

            \textbf{Dataset} & \textbf{Model} & \textbf{$Base$} & \textbf{$Q_{8}$} & \textbf{$ToMe$} & \textbf{$\gT_{intra}^{0.5}$} & \textbf{$\gT_{inter}^{100}$} & \textbf{$\gT_{inter}^{1000}$} & \textbf{$\gT_{inter}^{10000}$} \\

            \midrule

            Waterbirds & \multirow{5}{*}{CLIP}      & 123.99 & 73.89 & 91.64 & 110.93 & 102.65 & 107.93 & 117.36      \\
            CelebA &                                & 126.53 & 74.47 & 93.07 & 114.37 & 96.31 & 102.12 & 115.46      \\
            MetaShift &                             & 123.30 & 74.02 & 90.96 & 113.84 & 98.98 & 105.78 & 114.70      \\
            OpenImages (Com.) &                     & 135.05 & 74.80 & 94.63 & 114.62 & 106.33 & 112.67 & 123.95      \\
            OpenImages (Rare) &                     & 135.66 & 75.27 & 94.04 & 115.78 & 104.73 & 110.75 & 123.05     \\

            \midrule

            COCO & \multirow{2}{*}{BLIP}            & 191.74 & 81.68 & 147.23 & 169.77 & 144.71 & 156.25 & 171.84     \\
            NoCaps &                                & 185.12 & 81.89 & 149.26 & 175.06 & 141.20 & 155.11 & 170.38     \\

            \bottomrule
        \end{tabular}

    \caption{Power $\downarrow$}
    \end{subtable}

    \hspace{1em}

    \begin{subtable}{1\textwidth}
    \centering

        \begin{tabular}{r|c|c|cc|c|ccc}
            \toprule

            \textbf{Dataset} & \textbf{Model} & \textbf{$Base$} & \textbf{$Q_{8}$} & \textbf{$ToMe$} & \textbf{$\gT_{intra}^{0.5}$} & \textbf{$\gT_{inter}^{100}$} & \textbf{$\gT_{inter}^{1000}$} & \textbf{$\gT_{inter}^{10000}$} \\

            \midrule

            Waterbirds & \multirow{5}{*}{CLIP}      & 8.48 & 88.55 & 17.04 & 5.84 & 9.88 & 9.73 & 9.91       \\
            CelebA &                                & 8.33 & 89.41 & 17.25 & 5.86 & 9.45 & 9.54 & 9.99       \\
            MetaShift &                             & 8.21 & 89.48 & 17.23 & 6.01 & 9.99 & 9.87 & 9.75       \\
            OpenImages (Com.) &                     & 8.83 & 92.23 & 17.44 & 5.97 & 9.64 & 9.69 & 10.23       \\
            OpenImages (Rare) &                     & 8.60 & 86.00 & 17.04 & 6.19 & 9.69 & 10.02 & 9.67       \\

            \midrule

            COCO & \multirow{2}{*}{BLIP}            & 12.45 & 57.11 & 15.06 & 8.10 & 8.73 & 9.10 & 10.31       \\
            NoCaps &                                & 12.44 & 57.61 & 15.08 & 8.04 & 9.49 & 9.28 & 10.51       \\

            \bottomrule
        \end{tabular}

    \caption{Runtime $\downarrow$}
    \end{subtable}

    \caption{
        Power (watt) and runtime (millisecond) per sample. 
        The runtime is first converted to seconds before taking its product with the power to produce the energy values in Table \ref{tab:efficiency}.
    }
    \label{tab:power_runtime}
\end{table}

\begin{table}[!ht]
    \begin{subtable}{1\textwidth}
    \centering

        \begin{tabular}{r|c|c|c|cc|cc|cc}
            \toprule

            \textbf{Dataset} & \textbf{Model} & \textbf{$\gT_{intra}^{0.25}$} & \textbf{$\gT_{intra}^{0.33}$} & \textbf{$\gT_{intra}^{0.7}$}    \\

            \midrule

            Waterbirds & \multirow{5}{*}{CLIP}      & \multirow{5}{*}{148} & \multirow{5}{*}{132} & \multirow{5}{*}{59} \\
            CelebA & & & &  \\
            MetaShift & & & &  \\
            OpenImages (Com.) & & & &  \\
            OpenImages (Rare) & & & &  \\

            \midrule

            COCO & \multirow{2}{*}{BLIP}            & \multirow{2}{*}{433} & \multirow{2}{*}{386} & \multirow{2}{*}{173}    \\
            NoCaps & & & &  \\

            \bottomrule
        \end{tabular}

    \caption{Length}
    \end{subtable}

    \hspace{1em}

    \begin{subtable}{1\textwidth}
    \centering
    \resizebox{1\textwidth}{!}{%
        \begin{tabular}{c|cc|cc|cc|cc}
            \toprule

            \multirow{2}{*}{\textbf{Model}} & \multicolumn{2}{c|}{\textbf{Waterbirds}} & \multicolumn{2}{c|}{\textbf{CelebA}} & \multicolumn{2}{c}{\textbf{MetaShift}}      & \multicolumn{2}{c}{\textbf{OpenImages}}   \\

            & \textbf{Average} $\uparrow$ & \textbf{Worst} $\uparrow$      & \textbf{Average} $\uparrow$ & \textbf{Worst} $\uparrow$        & \textbf{Average} $\uparrow$ & \textbf{Worst} $\uparrow$    & \textbf{Common} $\uparrow$ & \textbf{Rare} $\uparrow$   \\

            \midrule

            $\gT_{intra}^{0.25}$        & 78.31 & 26.69     & 89.46 & 48.47     & 94.70 & 87.69     & 69.79 & 63.37     \\[1mm]
            $\gT_{intra}^{0.33}$        & 77.72 & 27.78     & 89.31 & 48.83     & 94.62 & 87.18     & 69.11 & 62.75     \\[1mm]
            $\gT_{intra}^{0.7}$         & 73.27 & 18.80     & 89.19 & 45.37     & 82.61 & 63.59     & 46.83 & 42.68     \\
    
            \bottomrule
        \end{tabular}
    }
    \caption{CLIP (image classification and subgroup robustness)}
    \end{subtable}

    \hspace{1em}

    \begin{subtable}{1\textwidth}
    \centering
    \resizebox{1\textwidth}{!}{%
        \begin{tabular}{c|cc|cccccccc}
            \toprule

            \multirow{3}{*}{\textbf{Model}}     & \multicolumn{2}{c|}{\textbf{COCO}}     & \multicolumn{8}{c}{\textbf{NoCaps}}  \\

            & \multicolumn{2}{c|}{\textbf{Karpathy} $\uparrow$} & \multicolumn{2}{c}{\textbf{In-Domain} $\uparrow$}       & \multicolumn{2}{c}{\textbf{Near-Domain} $\uparrow$}     & \multicolumn{2}{c}{\textbf{Out-of-Domain} $\uparrow$}      & \multicolumn{2}{c}{\textbf{Overall} $\uparrow$}    \\

            & BLEU@4 & CIDEr        & CIDEr & SPICE     & CIDEr & SPICE     & CIDEr & SPICE     & CIDEr & SPICE    \\

            \midrule

            $\gT_{intra}^{0.25}$       & 33.92 & 107.47        & 103.59 & 14.65        & 98.39 & 14.07        & 94.47 & 13.59     & 98.34 & 14.06     \\[1mm]
            $\gT_{intra}^{0.33}$       & 33.59 & 106.25        & 103.19 & 14.65        & 97.40 & 13.95        & 92.66 & 13.41     & 97.27 & 13.95     \\[1mm]
            $\gT_{intra}^{0.7}$        & 29.64 & 93.20         & 89.89  & 13.53        & 85.04 & 12.82        & 80.81 & 12.31     & 84.88 & 12.82     \\
    
            \bottomrule
        \end{tabular}
    }
    \caption{BLIP (image captioning)}
    \end{subtable}

    \caption{
        Token length (including [CLS]) and performance of the \emph{intra}-image approach with varying dropping ratios.
        Performance degrades naturally as the ratio increases before falling sharply at 0.7.
    }
    \label{tab:drop}
\end{table}

\begin{table}[!ht]
    \begin{subtable}{1\textwidth}
    \centering
    \resizebox{1\textwidth}{!}{%
        \begin{tabular}{c|cc|cc|cc|cc}
            \toprule

            \multirow{2}{*}{\textbf{Model}} & \multicolumn{2}{c|}{\textbf{Waterbirds}} & \multicolumn{2}{c|}{\textbf{CelebA}} & \multicolumn{2}{c}{\textbf{MetaShift}}      & \multicolumn{2}{c}{\textbf{OpenImages}}   \\

            & \textbf{Average} $\uparrow$ & \textbf{Worst} $\uparrow$      & \textbf{Average} $\uparrow$ & \textbf{Worst} $\uparrow$        & \textbf{Average} $\uparrow$ & \textbf{Worst} $\uparrow$    & \textbf{Common} $\uparrow$ & \textbf{Rare} $\uparrow$   \\

            \midrule

            $\gT_{intra}^{0.25}$        & 76.90 & 21.50     & 89.88 & 48.75     & 94.39 & 85.13     & 70.00 & 63.38     \\[1mm]
            $\gT_{intra}^{0.33}$        & 75.72 & 20.51     & 90.10 & 48.89     & 93.44 & 84.62     & 69.66 & 63.06     \\[1mm]
            $\gT_{intra}^{0.5}$         & 72.38 & 19.52     & 90.56 & 51.85     & 91.61 & 81.93     & 67.59 & 61.61     \\[1mm]
            $\gT_{intra}^{0.7}$         & 66.78 & 11.58     & 90.24 & 61.48     & 84.29 & 69.73     & 52.72 & 49.86     \\
    
            \bottomrule
        \end{tabular}
    }
    \caption{CLIP (image classification and subgroup robustness)}
    \end{subtable}

    \hspace{1em}

    \begin{subtable}{1\textwidth}
    \centering
    \resizebox{1\textwidth}{!}{%
        \begin{tabular}{c|cc|cccccccc}
            \toprule

            \multirow{3}{*}{\textbf{Model}}     & \multicolumn{2}{c|}{\textbf{COCO}}     & \multicolumn{8}{c}{\textbf{NoCaps}}  \\

            & \multicolumn{2}{c|}{\textbf{Karpathy} $\uparrow$} & \multicolumn{2}{c}{\textbf{In-Domain} $\uparrow$}       & \multicolumn{2}{c}{\textbf{Near-Domain} $\uparrow$}     & \multicolumn{2}{c}{\textbf{Out-of-Domain} $\uparrow$}      & \multicolumn{2}{c}{\textbf{Overall} $\uparrow$}    \\

            & BLEU@4 & CIDEr        & CIDEr & SPICE     & CIDEr & SPICE     & CIDEr & SPICE     & CIDEr & SPICE    \\

            \midrule

            $\gT_{intra}^{0.25}$        & 33.89 & 106.59        & 103.70 & 14.39        & 98.02 & 14.01        & 93.79 & 13.24     & 97.98 & 13.92     \\[1mm]
            $\gT_{intra}^{0.33}$        & 33.78 & 106.57        & 102.44 & 14.35        & 98.15 & 13.98        & 93.04 & 13.21     & 97.73 & 13.88     \\[1mm]
            $\gT_{intra}^{0.5}$         & 33.32 & 104.65        & 100.70 & 14.06        & 95.11 & 13.72        & 92.25 & 13.22     & 95.33 & 13.67     \\[1mm]
            $\gT_{intra}^{0.7}$         & 31.15  & 97.18        & 95.02  & 13.57        & 86.83 & 13.01        & 86.43 & 12.52     & 87.93 & 13.00     \\
    
            \bottomrule
        \end{tabular}
    }
    \caption{BLIP (image captioning)}
    \end{subtable}

    \caption{
        Token length (including [CLS]) and performance of the \emph{intra}-image approach with random dropping. 
        Unlike CelebA, the subgroup robustness degrades noticeably on Waterbirds and MetaShift. 
        Performance is also slightly lower on OpenImages but equivalent on the remaining datasets except for $\gT_{intra}^{0.7}$.
    }
    \label{tab:random_intra}
\end{table}

\begin{table}[!ht]
    \centering
    \resizebox{\textwidth}{!}{%
        \begin{tabular}{c|c|cc|cc|cc}
            \toprule

            \multirow{2}{*}{\textbf{Model}} & \multirow{2}{*}{\textbf{Subgroup}} & \multicolumn{2}{c|}{\textbf{Waterbirds}} & \multicolumn{2}{c|}{\textbf{CelebA}} & \multicolumn{2}{c}{\textbf{MetaShift}}   \\

            & & \textbf{Length} & \textbf{$\Delta$ Accuracy}      & \textbf{Length} & \textbf{$\Delta$ Accuracy}        & \textbf{Length} & \textbf{$\Delta$ Accuracy}  \\

            \midrule

            \multirow{4}{*}{$Base$}             & 0     & \multirow{4}{*}{197} & 98.89      & \multirow{4}{*}{197} & 95.63      & \multirow{4}{*}{197} & 98.54    \\
                                                & 1     &                      & 82.45      &                      & 96.23      &                      & 93.19    \\
                                                & 2     &                      & 21.86      &                      & 48.67      &                      & 87.69    \\
                                                & 3     &                      & 54.73      &                      & 50.00      &                      & 95.36    \\

            \midrule

            \multirow{4}{*}{$\gT_{inter}^{100}$}          & 0     & 144 & \textcolor{Red}{-0.67}      & 85 & \textcolor{Red}{-3.25}       & 118 & \textcolor{Red}{-0.74}      \\
                                                & 1     & 106 & \textcolor{Red}{-3.82}      & 88 & \textcolor{Red}{-0.13}       & 110 & \textcolor{Red}{-6.18}      \\
                                                & 2     & 140 & \textcolor{Green}{19.00}    & 98 & \textcolor{Green}{32.62}     & 119 & \textcolor{Red}{-2.92}      \\
                                                & 3     & 105 & \textcolor{Green}{6.17}     & 99 & \textcolor{Green}{6.30}      & 100 & \textcolor{Red}{-9.02}      \\

            \midrule

            \multirow{4}{*}{$\gT_{inter}^{1000}$}         & 0     & 160 & \textcolor{Red}{-0.40}      & 119 & \textcolor{Red}{-0.46}      & 138 & \textcolor{Red}{-0.25}      \\
                                                & 1     & 129 & \textcolor{Red}{-0.22}      & 117 & \textcolor{Green}{0.74}     & 138 & \textcolor{Red}{-2.06}      \\
                                                & 2     & 160 & \textcolor{Green}{9.03}     & 124 & \textcolor{Green}{18.92}    & 143 & \textcolor{Red}{-1.17}      \\
                                                & 3     & 129 & \textcolor{Green}{2.18}     & 124 & \textcolor{Red}{-4.07}      & 131 & \textcolor{Red}{-2.13}      \\

            \midrule

            \multirow{4}{*}{$\gT_{inter}^{10000}$}        & 0     & 180 & \textcolor{Red}{-0.09}      & 158 & \textcolor{Green}{0.25}     & 163 & \textcolor{Red}{-0.37}      \\
                                                & 1     & 156 & \textcolor{Green}{1.34}     & 151 & \textcolor{Green}{0.66}     & 167 & \textcolor{Green}{0.19}     \\
                                                & 2     & 181 & \textcolor{Green}{4.75}     & 158 & \textcolor{Green}{2.96}     & 169 & \textcolor{Red}{-1.75}      \\
                                                & 3     & 158 & \textcolor{Green}{1.71}     & 156 & \textcolor{Red}{-6.30}      & 163 & \textcolor{Red}{-0.81}      \\

            \bottomrule
        \end{tabular}
    }
    \caption{
        Distribution of token length (including [CLS]) and accuracy (w.r.t. $Base$) by subgroup. 
        $\gT_{inter}^{\gV}$ of varying vocabulary sizes are shown. 
        Like text tokenizers, VWTs may induce unequal token lengths as seen with $\gT_{inter}^{100}$ on Waterbirds. 
        Performance is also affected unequally as a stronger sequence compression does not correlate with a greater improvement or degradation in accuracy.
    }
    \label{tab:fairness}
\end{table}

\end{document}